%% file: FinixDoc_main.tex
\documentclass[dvipsnames]{article}
\usepackage{style/colm2024_conference}

\usepackage{xeCJK}
\setCJKsansfont[
  BoldFont=FandolHei-Bold.otf
]{FandolHei-Regular.otf}
\setCJKmonofont{FandolFang-Regular.otf}

\usepackage{needspace}
\usepackage{booktabs}
\usepackage{changepage}
\usepackage{graphicx}
\usepackage{enumitem}
\usepackage{wrapfig}
\usepackage{float}
\usepackage{algorithm}
\usepackage{algpseudocode}
\usepackage{microtype}
\usepackage{amsmath}
\usepackage{colortbl}
\definecolor{lightgray}{rgb}{0.9,0.9,0.9}
\usepackage{caption}
\usepackage{subcaption}
\usepackage{setspace}
\usepackage{multirow}
\usepackage{tabularx}
\usepackage{pgfplots}
\pgfplotsset{compat=1.18}
\usepackage{tikz}
\usetikzlibrary{er,positioning,bayesnet}
\usepackage{makecell}
\usepackage{tipa}
\usepackage{siunitx}
\usepackage{nicefrac}
\usepackage{tocloft}
\usepackage{listings}
\usepackage[raster,skins]{tcolorbox}
\usepackage{xltabular}
\usepackage{adjustbox}
\usepackage{xurl}
\usepackage{rotating}
\usepackage[misc]{ifsym}
\usepackage[normalem]{ulem}
\useunder{\uline}{\ul}{}

\input{tool/math_commands.tex}

\newcommand*\justify{%
  \fontdimen2\font=0.4em
  \fontdimen3\font=0.2em
  \fontdimen4\font=0.1em
  \fontdimen7\font=0.1em
  \hyphenchar\font=`\-
}

\renewcommand{\texttt}[1]{%
  \begingroup
  \ttfamily
  \begingroup\lccode`~=`/\lowercase{\endgroup\def~}{/\discretionary{}{}{}}%
  \begingroup\lccode`~=`[\lowercase{\endgroup\def~}{[\discretionary{}{}{}}%
  \begingroup\lccode`~=`.\lowercase{\endgroup\def~}{.\discretionary{}{}{}}%
  \catcode`/=\active\catcode`[=\active\catcode`.=\active
  \justify\scantokens{#1\noexpand}%
  \endgroup
}

\title{FinixDoc: Rethinking Financial Document Parsing\\
Beyond Saturated Benchmarks}

\author{
  \textbf{Hang Wang}\textsuperscript{*},
  \textbf{Jin Zhang}\textsuperscript{*},
  \textbf{Guoliang Xu}\textsuperscript{*},
  \textbf{Pengyue Lu} \\
  \textbf{Yao Li},
  \textbf{Zijiao Zhang},
  \textbf{Tianyu Huang},
  \textbf{Weiqi Xiong},
  \textbf{Yulong Wang},
  \textbf{Chuqiao Lu} \\
  \textbf{Wenkang Huang},
  \textbf{Kai Yang},
  \textbf{Yadong Li},
  \textbf{Hui Li},
  \textbf{Xingzhong Xu}\textsuperscript{\ensuremath{\ddagger}},
  \textbf{Xiao Xu}\textsuperscript{\Letter} \\
  \textsuperscript{*}Equal contribution.
  \quad
  \textsuperscript{\ensuremath{\ddagger}}Team leader.
  \quad
  \textsuperscript{\Letter}Corresponding author. \\
  Ant Group, Hangzhou, China \\
  \mbox{\Letter\ \texttt{suyan.xx@antgroup.com}} \\
}

\begin{document}

\maketitle

\begin{abstract}
Financial document parsing requires accuracy, structural consistency, and verifiability that current benchmarks often fail to reflect. We present \textbf{FinixDoc}, an end-to-end agentic parsing system for real-world financial documents, with \textbf{FinixDoc-VL}, a 4B-scale vision-language model built on Qwen3-VL-4B, as its core parser. To characterize the gap between benchmark and deployment performance, we introduce a \textbf{Document Parsing Capability Matrix} organized along two practical axes: visual quality and document scale. Guided by this matrix, FinixDoc-VL is trained with a domain-adapted recipe combining homoglyph-aware contrastive learning and multi-stage reinforcement learning with composite domain-specific rewards. To better leverage our accumulated advantage in low-quality financial-document data and support large-scale, high-quality data production, we further build a human-in-the-loop \textbf{Data Factory} pipeline with confidence-aware expert review. For evaluation, we construct \textbf{FinixDocBench}, a financial-domain evaluation suite covering digital-native, camera-captured, ultra-large-page, and internal-workflow scenarios, with a compliance-reviewed subset released alongside this technical report. On its main subsets, FinixDoc-VL achieves the highest overall score (\textbf{81.43}) among evaluated baselines, outperforming the next-best open-source model by 5.13 points, with the largest gains on internal financial workflows (\textbf{FinixInner: 84.08} vs.\ 78.73).
\end{abstract}

\vspace{-1.4em}
\enlargethispage{1.2cm}

\noindent\begin{minipage}{\textwidth}
\centering
\includegraphics[
    width=0.88\textwidth,
    height=0.4\textheight,
    keepaspectratio,
    trim=0 0 0 20,
    clip
]{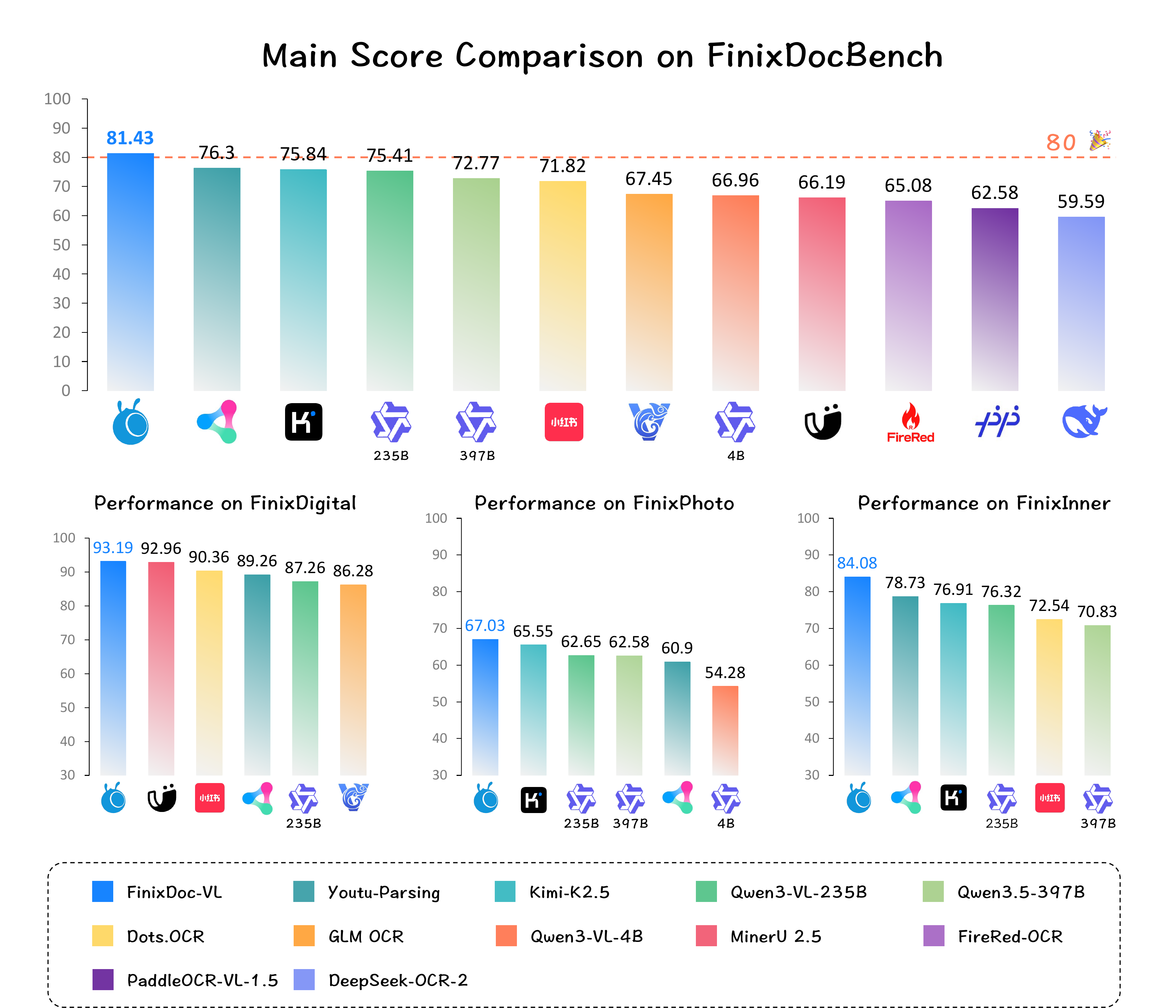}

\captionof{figure}{\small
\textbf{Main score comparison on FinixDocBench.}
The main score is computed as the macro-average over FinixDigital, FinixPhoto, and FinixInner.
The top panel reports the overall main score, and the bottom panels show subset-level performance.
FinixHuge is excluded from this comparison and reported separately, since it evaluates ultra-large-page documents under a system-level protocol with limited model coverage.
}
\label{fig:finixdocbench-main-score}
\end{minipage}

\vfill
\newpage

\input{content/1_introduction}
\input{content/2_FinixDoc}
\input{content/3_dataset}
\input{content/4_evaluation}
\input{content/5_related_work}
\input{content/6_conclusion}

\clearpage
\bibliography{tool/biblio}
\bibliographystyle{style/colm2024_conference}

\clearpage
\appendix
\input{content/appendix}

\end{document}

%% file: tool/math_commands.tex
\usepackage{amsmath,amsfonts,bm}

\def\eqref#1{equation~\ref{#1}}

\def\1{\bm{1}}

\DeclareMathAlphabet{\mathsfit}{\encodingdefault}{\sfdefault}{m}{sl}
\SetMathAlphabet{\mathsfit}{bold}{\encodingdefault}{\sfdefault}{bx}{n}



%% file: content/1_introduction.tex
\section{Introduction}

As a financial technology enterprise operating large-scale financial workflows, we process massive volumes of heterogeneous financial documents every day, including insurance policies, medical receipts, claim materials, identity documents, expense statements, and financial reports. These documents arrive in diverse forms, such as digitally native PDFs, scanned copies, office files, and camera-captured images, and their parsed contents directly support downstream business processes that require high accuracy, structural consistency, traceability, and verifiability. This real-world industrial setting motivates our focus on financial document parsing.

Document parsing is therefore a foundational capability for document digitization, content understanding, and information extraction. Traditional approaches typically rely on task-specific pipelines composed of optical character recognition (OCR), layout analysis, and field extraction \citep{smith2007tesseract,zhang2024documentparsingunveiled,paddleocr2025technical}. Although these methods can perform well in scenarios with well-defined rules, they usually face limited performance ceilings, high maintenance costs, and weak generalization across templates and application settings.

In recent years, Vision-Language Models (VLMs) have driven a major paradigm shift in document understanding \citep{kim2022donut,lee2023pix2struct,blecher2023nougat,bai2025qwen25vl}. Unlike traditional pipelines that decouple recognition, structural analysis, and field extraction, modern document parsing is increasingly treated as a unified modeling problem \citep{zhang2024documentparsingunveiled,kim2022donut,cui2025paddleocrvl,dotsocr2025}. In this work, we use the term \textit{parsing} to refer to this unified capability, which jointly encompasses text recognition, layout detection and categorization, table structure recovery, and reading-order reconstruction, with the final output rendered as either structured JSON or full-page Markdown.

In large-scale financial workflows, document parsing is not merely a generic OCR or layout-analysis task. Its outputs directly affect business processes and therefore impose stricter requirements than standard benchmark settings. These requirements expose failure modes that are often underrepresented in standard document benchmarks.

In real-world financial deployments, we observe a clear gap between benchmark performance and practical utility. Many models that achieve strong results on standard benchmarks (e.g., OmniDocBench \citep{ouyang2025omnidocbench}) still exhibit major shortcomings in production settings, including weak robustness, structural inconsistencies, and failures on long documents or large images. Such behavior falls well short of the stringent financial requirements for accuracy, consistency, and verifiability.

\begin{figure}[t!]
  \centering
  \scalebox{1}[0.9]{%
    \includegraphics[width=\linewidth]{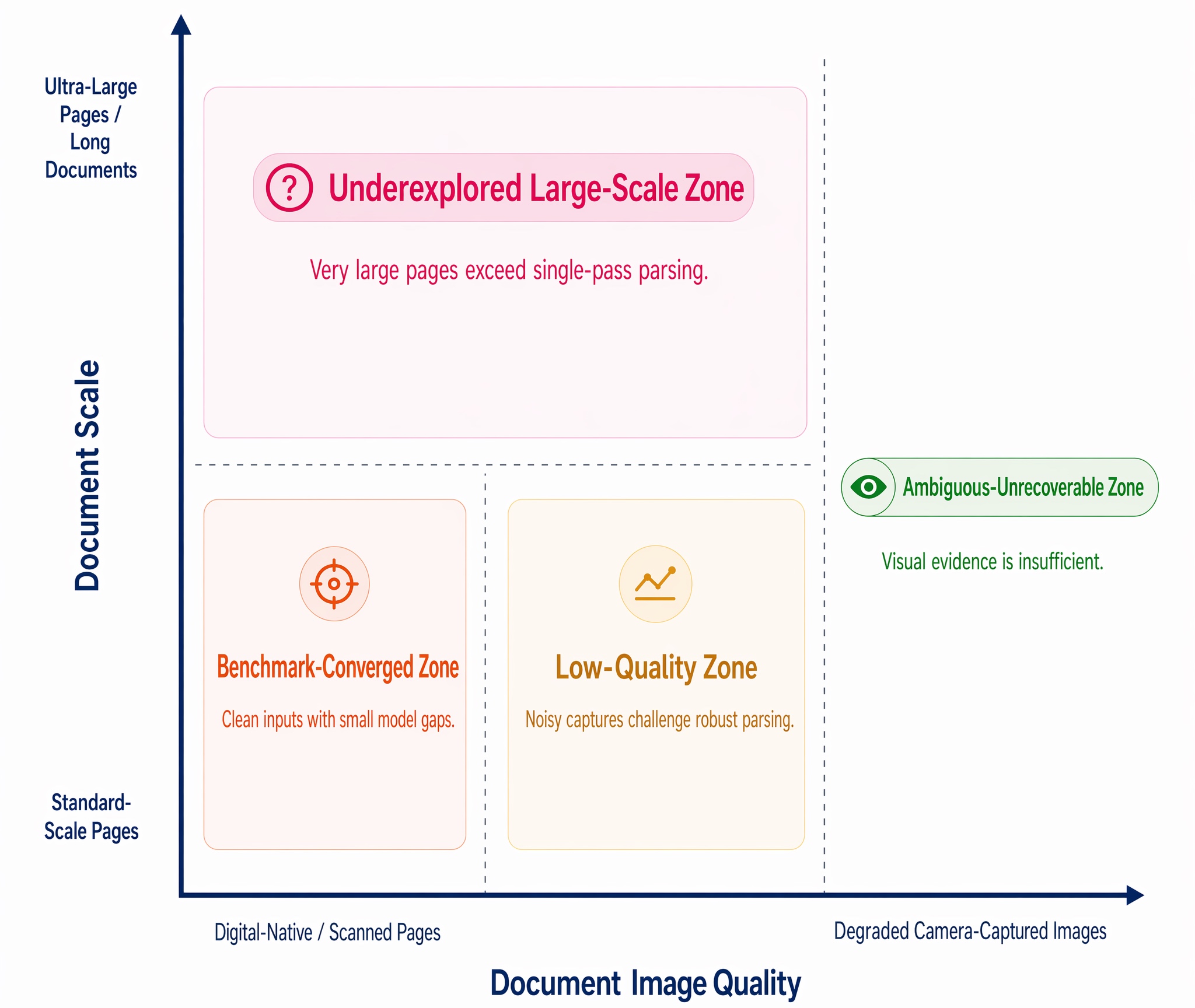}%
  }
  \caption{\textbf{Document Parsing Capability Matrix.} The horizontal axis represents decreasing document quality, ranging from digital or scanned documents to camera-captured documents affected by perspective distortion, creases, and blur. The vertical axis represents increasing document scale, ranging from single-page or single-image inputs to ultra-long documents and ultra-large images. Based on this matrix, document parsing scenarios can be categorized into the Benchmark-Converged Zone, Low-Quality Zone, Underexplored Large-Scale Zone, and Ambiguous-Unrecoverable Zone.}
  \label{fig:capability-matrix}
\end{figure}

Through extensive analysis of failure cases in real financial deployments, we observe that these shortcomings are not isolated incidents but instead vary systematically along two orthogonal dimensions: the visual quality of the input document and the scale of the page or document. Accordingly, we introduce a \textit{Document Parsing Capability Matrix} from the perspective of financial document processing. As illustrated in Figure~\ref{fig:capability-matrix}, this matrix organizes document parsing scenarios along two dimensions: document quality and document scale.

Based on this matrix, we organize current document parsing scenarios into several representative capability zones. First, mainstream benchmarks are concentrated largely in the bottom-left region, which we term the \textbf{Benchmark-Converged Zone}. Documents in this region typically have clear page quality, regular layouts, sufficiently high resolution, and manageable scale. As a result, many VLMs, especially document-specific models, already achieve strong performance in this zone, and the differences among leading models are often small.

Second, along the horizontal axis, we define the \textbf{Low-Quality Zone}, which corresponds to scenarios with degraded document quality. This zone includes documents commonly encountered in real-world financial workflows, such as medical receipts, claim materials, identity cards, and reimbursement vouchers. Because these documents are often captured with mobile phones, they are frequently affected by blur, shadows, glare, creases, perspective distortion, and partial occlusion. In this zone, we observe substantial performance degradation across most existing solutions. More importantly, document-specific VLMs that perform strongly in the Benchmark-Converged Zone often underperform strong general-purpose VLMs under these conditions. We hypothesize that, compared with general-purpose VLMs, specialized document models typically have smaller parameter scales, narrower training objectives, and more limited data diversity, making them more sensitive to distribution shifts and less robust in challenging real-world conditions.

Third, in the upper region of the matrix, we define the \textbf{Underexplored Large-Scale Zone}, which corresponds to large-scale documents that remain insufficiently covered by existing benchmarks and systems. Representative examples include ultra-long insurance policies and densely structured multi-column tables. When a document becomes excessively large (e.g., over 100 million pixels), directly feeding the entire page into a single VLM poses a major challenge for most existing models. This difficulty arises either because aggressive downsampling removes critical local details, or because the resulting input and output token sequences exceed practical model capacity. We argue that this challenge is unlikely to be solved by simply extending the context window of VLMs. Instead, it requires a system-level solution that jointly addresses local content perception, global structural modeling, and cross-region result integration. In this paper, our primary focus is the Low-Quality Zone. For the Underexplored Large-Scale Zone, we present a practical split-then-merge strategy as an initial solution; its design and implementation details are deferred to Appendix~\ref{appendix:tiling_merging}.

Finally, on the far right of the matrix, we identify the \textbf{Ambiguous-Unrecoverable Zone}. When document quality becomes extremely poor---beyond reliable visual recognition---current VLMs tend to rely on language priors and contextual regularities to generate speculative predictions \citep{li2023evaluatingpope,liu2024hallucinationsurvey}. The resulting outputs may appear plausible, yet lack factual grounding in the image itself. While such behavior may occasionally be tolerated in open-ended question answering, it is unacceptable in financial document parsing. Therefore, suppressing hallucinations under low-confidence conditions and adhering to the industrial principle of ``better omission than error'' remain central challenges for future financial document parsing, especially in risk-sensitive settings where precision is preferred over recall.

To address these challenges, we propose \textbf{FinixDoc}, an end-to-end agentic parsing system tailored to financial documents, with \textbf{FinixDoc-VL} as its core vision-language model. We build FinixDoc-VL on top of the Qwen3-VL-4B architecture \citep{qwen2025qwen3vl}: as discussed above, general-purpose VLMs exhibit stronger robustness than document-specific models in the Low-Quality Zone, while a 4B-scale model offers a practical balance between this robustness and the latency, memory, and cost constraints of real financial deployment. Starting from this backbone, the design of FinixDoc is explicitly organized around the capability zones identified above. For the \textit{Low-Quality Zone}, which is the primary focus of this work, we adapt the visual encoder to financial glyphs through homoglyph-based contrastive learning \citep{radford2021learning,khosla2020supervised}, and align the model's outputs with business-level requirements through reinforcement learning \citep{shao2024deepseekmath,guo2025deepseekr1} with composite domain-specific rewards; both stages are further supported by an increased proportion of real-world camera-captured financial documents in training. For the \textit{Underexplored Large-Scale Zone}, we introduce a system-level split-then-merge strategy in the Document Preprocessing Tool Layer, which decomposes ultra-large pages into tractable regions and reconstructs page-level outputs through direction-aware merging. For the \textit{Ambiguous-Unrecoverable Zone}, where confident-yet-ungrounded outputs are unacceptable in financial settings, we adopt the operational principle of ``better omission than error'' as a design constraint throughout the system, and leave the development of explicit uncertainty-aware refusal mechanisms as future work.

Empirically, FinixDoc-VL substantially improves over the Qwen3-VL-4B base model \citep{qwen2025qwen3vl} on OmniDocBench \citep{ouyang2025omnidocbench}, especially on the text and table metrics most relevant to financial documents. More importantly, on the financial-domain benchmark introduced in this work, FinixDoc-VL achieves state-of-the-art performance among major open-source models, with especially large gains in the Low-Quality and Underexplored Large-Scale Zones. These results suggest that progress in real-world financial document parsing should not be judged solely by scores on benchmarks drawn from the Benchmark-Converged Zone. Rather, it should be assessed by whether a system can robustly handle low-quality, high-complexity, and large-scale documents in practical settings.

The main contributions of this paper are summarized as follows:
\begin{itemize}
    \item We introduce a \textit{Document Parsing Capability Matrix} that characterizes real-world document parsing challenges along the dimensions of document quality and scale, providing a practical framework for analyzing industrial capability boundaries.
    \item We propose targeted training strategies for financial document parsing, including a ViT adaptation strategy for the financial visual domain and a business-aligned reinforcement learning optimization method, which substantially improve model performance in low-quality scenarios.
    \item We build an AI-driven \textit{Data Factory} pipeline that supports continuous data expansion and iterative refinement for financial document parsing.
    \item We introduce \textbf{FinixDocBench}, a benchmark closely aligned with real-world financial deployment scenarios, enabling systematic evaluation across digital-native, camera-captured, ultra-large-page, and internal workflow settings. A compliance-reviewed subset of FinixDocBench is released alongside this technical report to support broader research.
\end{itemize}

%% file: content/2_FinixDoc.tex
\section{FinixDoc}

\subsection{Overall Architecture}
\label{sec:overall_architecture}

FinixDoc is an end-to-end \textbf{agentic parsing system} for financial document parsing \citep{yao2023react,schick2023toolformer}. Given an input document, the agent leverages a core vision-language model, \emph{FinixDoc-VL}, alongside a suite of deterministic image preprocessing tools to transform raw inputs into robust, page-level visual representations. Rather than executing a static pipeline, the agent dynamically inspects each incoming document and conditionally invokes specific tools based on observed properties, such as input modality, page scale, and image quality indicators.

\begin{figure}[!t]
  \centering
  \includegraphics[width=\linewidth]{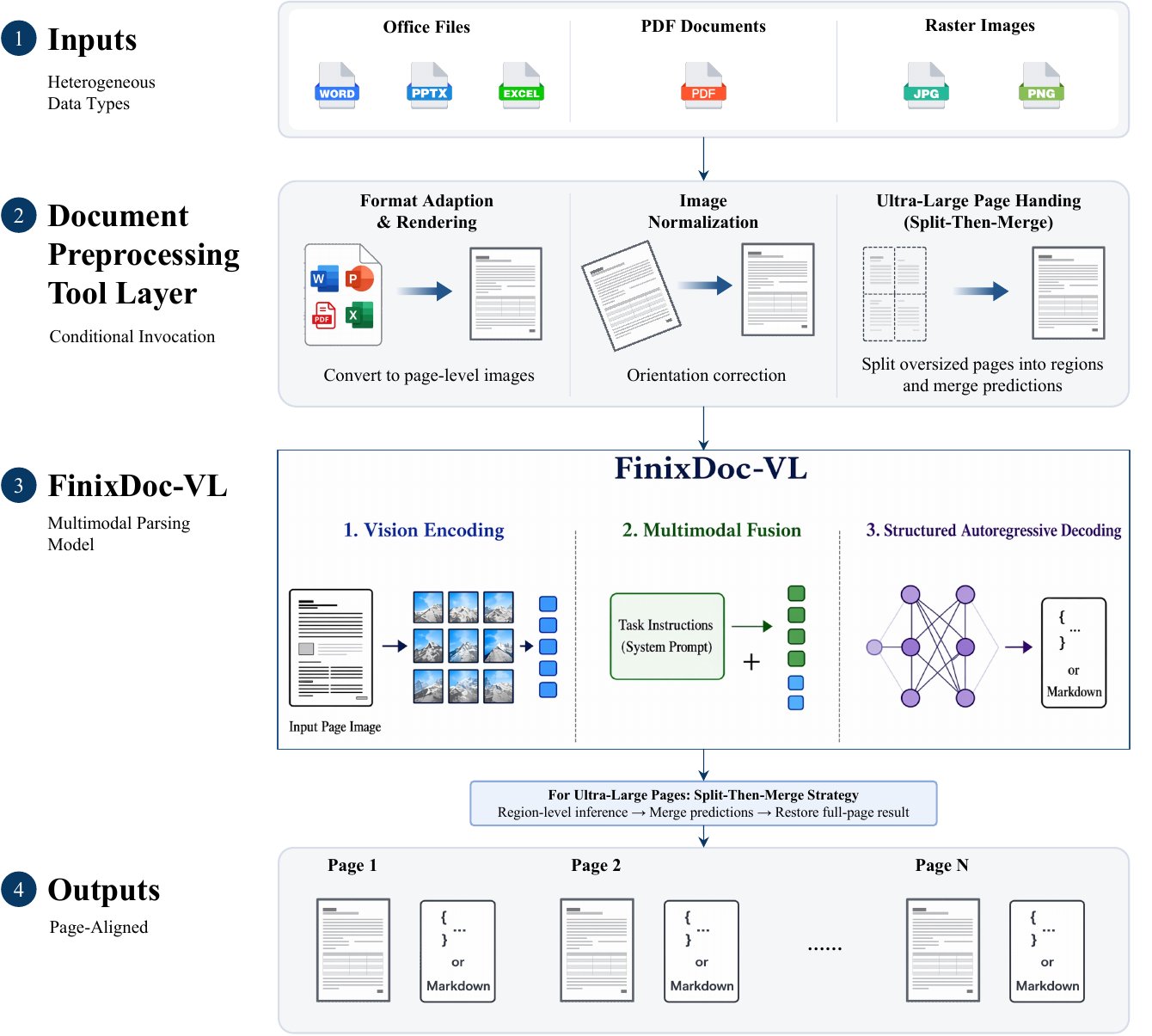}
  \caption{\textbf{Overall architecture of FinixDoc. The output format (JSON or Markdown) is specified by the user prompt.}}
  \label{fig:FinixDocAgent}
\end{figure}

\subsubsection*{Supported Inputs and Page-Aligned Outputs}

\begin{figure}[!t]
  \centering
  \includegraphics[width=\linewidth]{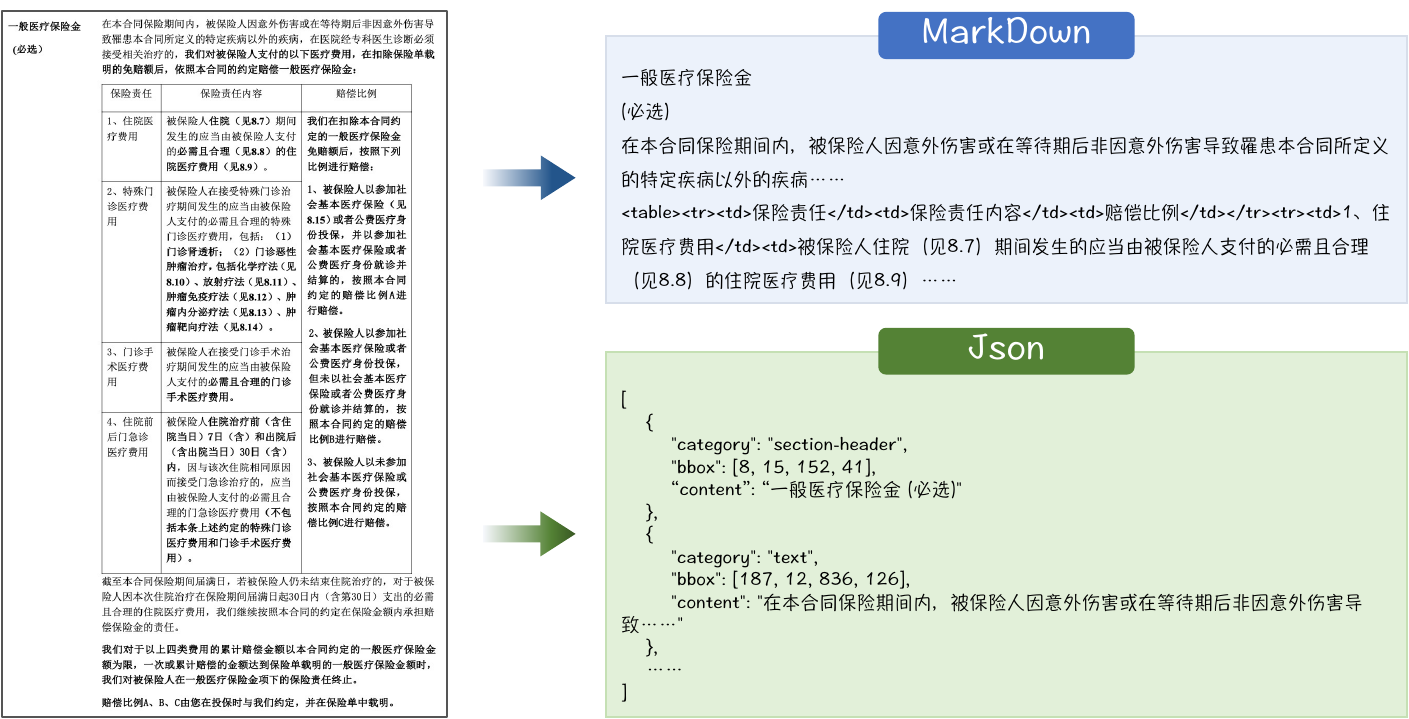}
  \caption{\textbf{Parsing output examples on \texttt{example.pdf}: Markdown (top) vs. JSON (bottom).}}
  \label{fig:parsing-example}
\end{figure}

FinixDoc accommodates heterogeneous data types, including office files (Word/PPT/Excel), PDFs, and raster images (PNG/JPG). Across all modalities, the final structured outputs (whether JSON or Markdown) are strictly \textbf{page-aligned}, ensuring that the number of output structures perfectly matches the source pages. The agent generates the specific output format (JSON vs. Markdown) according to the format requested in the user's prompt. An illustrative example of the two parsing output formats (JSON and Markdown) generated from \texttt{example.pdf} is shown in Fig.~\ref{fig:parsing-example}.

For ultra-large pages, the agent employs a \emph{split-then-merge} strategy: it partitions the oversized page into manageable regions for inference and subsequently stitches the region-level predictions back into a unified result that preserves the original full-page boundaries.

\subsubsection*{Document Preprocessing Tool Layer}

The tool layer provides practical utilities designed to enhance stability and robustness in complex real-world deployments. Typical tools include:
\begin{itemize}
\item \textbf{Format Adaptation and Rendering}: Converts office documents and PDFs into page-level images while decoding and normalizing native image inputs.
\item \textbf{Image Normalization}: Applies lightweight operations, such as scaling and orientation correction (e.g., deskewing or rotation), when quality degradation or misalignment is detected.
\item \textbf{Ultra-Large Page Handling}: Executes the \emph{split-then-merge} operation when a page exceeds standard visual scales, ensuring the core model processes appropriately sized regions without compromising layout fidelity.
\end{itemize}

Tool invocation is strictly \emph{conditional}: transformations are applied only when necessary for a specific instance, thereby avoiding unnecessary transformations that could degrade the original document layout. Concretely, image normalization is triggered when lightweight pre-checks detect orientation misalignment or severe scale anomalies, while ultra-large page handling is activated only when the input exceeds the effective visual token budget of FinixDoc-VL or exhibits structural complexity beyond reliable single-pass parsing. The detailed routing criteria for ultra-large page handling are described in Appendix~\ref{appendix:tiling_merging}, which covers token-load estimation and structural-complexity scoring.

\subsubsection*{FinixDoc-VL}
FinixDoc-VL serves as the multimodal core of the parsing system. Starting from the Qwen3-VL-4B backbone \citep{qwen2025qwen3vl}, FinixDoc-VL is fine-tuned for robust recognition, layout comprehension, and structured generation under low-quality, industrial conditions. Architecturally, it follows the standard VLM inference pipeline of \textbf{vision encoding} (mapping the input page image into visual tokens via the vision encoder, followed by spatial merging/compression), \textbf{multimodal fusion} (encoding task-specific instructions and fusing them with visual representations into a unified multimodal token sequence), and \textbf{structured autoregressive decoding} (generating the final outputs in Markdown or JSON conditioned on the fused representations) \citep{dosovitskiy2021image,bai2025qwen25vl}. FinixDoc-VL incorporates two domain-specific adaptations: visual representation adaptation in the encoder and business-aligned optimization in the decoder. Both are introduced through the training recipe described in Section~\ref{sec:training_recipe}, rather than through architectural modifications. This design enables FinixDoc to retain the end-to-end parsing capability of FinixDoc-VL while also providing practical advantages for real-world deployment. In particular, the system can robustly handle low-quality, camera-captured inputs and can be extended to process ultra-large pages using the \emph{split-then-merge} strategy described above.

\subsection{Training Recipe}
\label{sec:training_recipe}

As indicated by the Low-Quality Zone in the Capability Matrix, general-purpose vision-language models (VLMs) often exhibit stronger robustness than document-specific models when faced with real-world imaging artifacts such as blur, distortion, and shadows. In contrast, document-specific models are often trained on narrower data distributions and are therefore more vulnerable when the input deviates from their training assumptions. However, when adapting general-purpose VLMs to financial document parsing, we observe two major limitations:

\begin{enumerate}
    \item \textbf{Insufficient adaptation of visual representations to the financial domain.} \\
    The visual encoders used in general-purpose VLMs (e.g., Vision Transformers, ViTs \citep{dosovitskiy2021image}) are typically initialized or adapted from large-scale visual or vision-language pretraining \citep{radford2021learning,bai2025qwen25vl}. Real-world financial documents, however, differ substantially from these pre-training sources: they often exhibit denser table structures, stronger layout regularities, and more severe quality degradation introduced during image capture. This mismatch in both domain and difficulty makes it challenging for the model to capture the visual patterns and layout constraints that are specific to financial documents.

    \item \textbf{Mismatch between existing optimization objectives and financial task requirements.} \\
    Many document VLMs are still trained primarily with supervised fine-tuning (SFT). Although SFT helps the model imitate the output format and style of the training annotations, it often generalizes poorly to the highly variable and structurally complex cases encountered in financial workflows. Recent work has introduced reinforcement learning (RL) to alleviate some of the limitations of SFT, but the corresponding reward functions are usually designed for general-purpose tasks \citep{ouyang2022training,schulman2017proximal,shao2024deepseekmath,guo2025deepseekr1}. Although they address objectives such as instruction following and preference alignment, they do not adequately capture domain-critical information tied to financial semantics, layout structure, and high-risk fields.
\end{enumerate}

To address these issues, we train \textbf{FinixDoc-VL} using a two-stage optimization scheme tailored to financial document parsing. Starting from Qwen3-VL-4B as the base model, we retain the general robustness of a strong lightweight VLM while adapting it to the financial domain. In the first stage, we enhance visual and multimodal representations through contrastive learning to improve discrimination of fine-grained financial glyphs and structures. In the second stage, we apply reinforcement learning with domain-specific rewards so that optimization is more closely aligned with the practical requirements of financial document parsing.

\subsubsection{Financial Visual Representation Adaptation via Contrastive Learning}
\label{sec:contrastive_learning}

To mitigate the domain shift of general visual encoders in financial document scenarios, we propose a contrastive learning approach \citep{radford2021learning,khosla2020supervised} built on top of the visual branch of Qwen3-VL \citep{qwen2025qwen3vl}. The key idea is to construct high-quality hard negatives using visually similar characters (homoglyphs) centered around high-frequency characters in financial documents, and to use them to regularize both the visual representation space and the multimodal generative representation space. This improves the model's ability to discriminate fine-grained glyph differences that are particularly important in financial applications.

Unlike general image--text contrastive learning methods, our approach does not rely on large-scale weakly supervised image--text pairs. Instead, we construct triplets of \emph{(document image, correct text, hard negative text)} tailored to financial document parsing. Importantly, the hard negative texts are not produced by random perturbation. They are generated through homoglyph substitutions derived from real financial document errors, thereby better approximating the error patterns observed in deployment. This stage consists of three components:

\begin{enumerate}
    \item constructing a financial homoglyph vocabulary and generating hard negative samples accordingly;
    \item defining dual-path contrastive objectives over both visual and multimodal generative representations;
    \item jointly optimizing the contrastive objectives together with the supervised fine-tuning objective.
\end{enumerate}

\begin{figure}[t]
    \centering
    \includegraphics[width=0.95\linewidth]{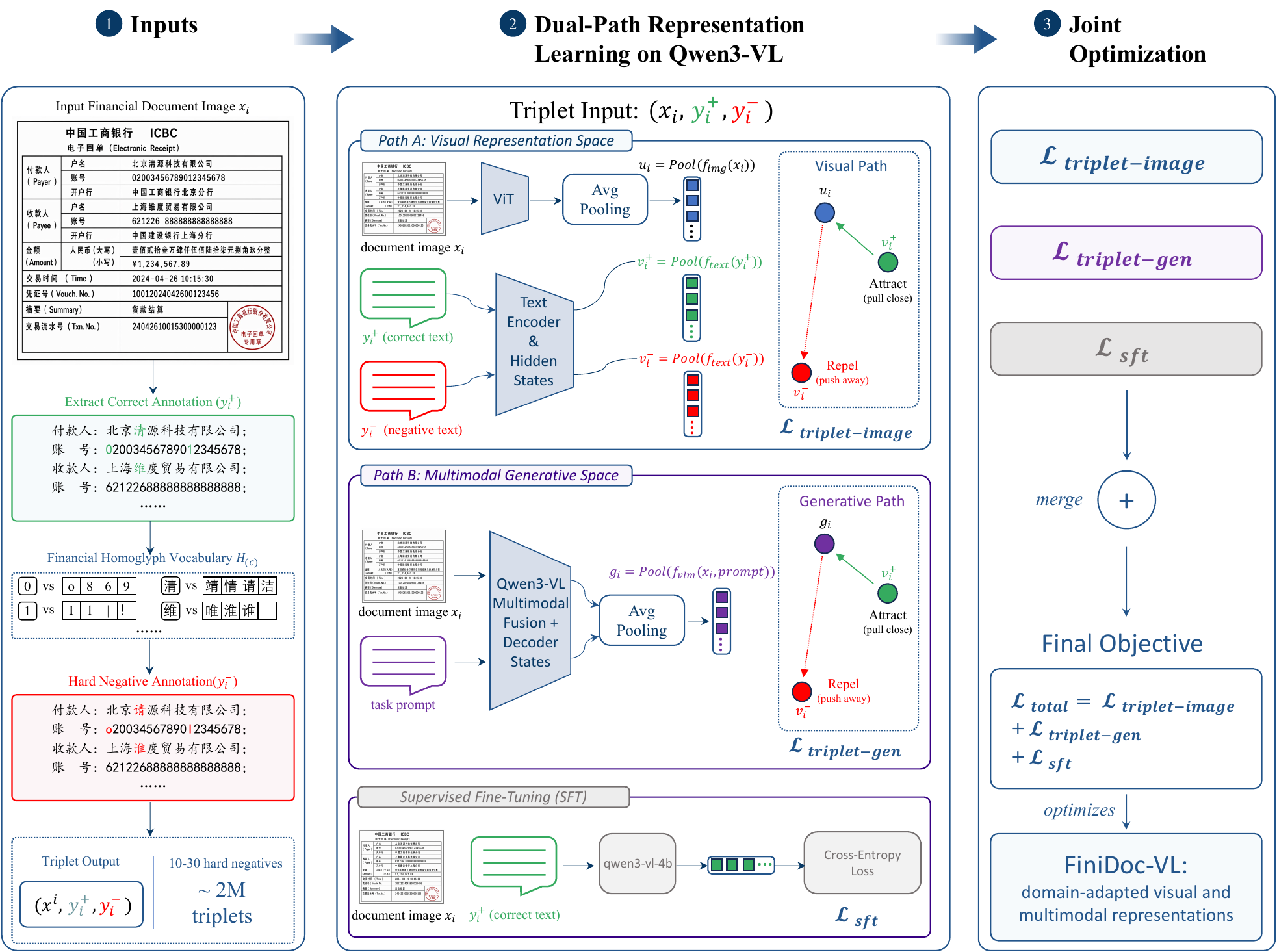}
    \caption{Overall training framework of the proposed financial visual representation adaptation method via contrastive learning. Given a document image and its ground-truth text, hard negative texts are generated through financial homoglyph substitution. The model is then optimized with dual-path contrastive objectives over both visual and multimodal generative representations, jointly with the supervised fine-tuning objective.}
    \label{fig:contrastive_learning}
\end{figure}

The overall training framework is illustrated in Figure~\ref{fig:contrastive_learning}.

\vspace{0.5em}
\noindent\textbf{Financial Homoglyph Vocabulary and Negative Sample Construction}
\vspace{0.5em}

We first analyze approximately 100,000 real financial documents and compile a candidate set of more than 4,500 high-frequency characters, including Chinese characters, English letters, digits, and common bilingual symbols. In addition, we build a stroke-order lexicon and a stroke-count lexicon covering more than 20,000 common Chinese characters to facilitate the retrieval of visually similar candidates.

To identify homoglyphs for the 4,500+ high-frequency characters, we design a glyph-similarity scoring module for estimating pairwise similarity between characters. Motivated by the main sources of visual confusion in real financial documents, we characterize character similarity from two major perspectives: rendered-glyph visual similarity and stroke-sequence similarity. The stroke-count signal is further used as a lightweight auxiliary cue during candidate retrieval, but is not used as an independent decisive score in the final filtering stage, since identical stroke counts alone do not necessarily imply visual similarity.

For rendered-glyph visual similarity, we render each character as a grayscale image of fixed size. Given two characters $c_i$ and $c_j$, with rendered images $I_i$ and $I_j$, respectively, we compute the structural similarity index \citep{wang2004image}:
\begin{equation}
S_{\mathrm{SSIM}}(c_i, c_j) = \mathrm{SSIM}(I_i, I_j).
\end{equation}
In addition, we compute perceptual hash similarity to capture coarse visual resemblance between rendered glyphs:
\begin{equation}
S_{\mathrm{pHash}}(c_i, c_j)
=
1 -
\frac{
\mathrm{Ham}\left(
\mathrm{pHash}(I_i), \mathrm{pHash}(I_j)
\right)
}{K},
\end{equation}
where $\mathrm{Ham}(\cdot,\cdot)$ denotes the Hamming distance and $K$ is the hash length. The rendered-glyph visual similarity is then defined as
\begin{equation}
S_{\mathrm{vis}}(c_i, c_j)
=
\max \left(
S_{\mathrm{SSIM}}(c_i, c_j),
S_{\mathrm{pHash}}(c_i, c_j)
\right).
\end{equation}

Rendered glyph images alone are insufficient to fully capture the internal composition of Chinese characters. We therefore introduce stroke-sequence similarity. Let the stroke-order sequences of characters $c_i$ and $c_j$ be
\begin{equation}
\mathbf{s}_i = (s_{i,1}, \dots, s_{i,m}), \qquad
\mathbf{s}_j = (s_{j,1}, \dots, s_{j,n}).
\end{equation}
We first compute a normalized edit-distance similarity \citep{levenshtein1966binary}:
\begin{equation}
S_{\mathrm{edit}}(c_i, c_j)
=
1 -
\frac{\mathrm{EditDist}(\mathbf{s}_i,\mathbf{s}_j)}{\max(m,n)}.
\end{equation}
To improve robustness to local order shifts and length mismatches, we further combine Dynamic Time Warping (DTW) \citep{sakoe1978dynamic} and Longest Common Subsequence (LCS) signals:
\begin{equation}
S_{\mathrm{order}}(c_i, c_j)
=
\sigma \left(
0.7 \left(1 - \frac{\mathrm{DTW}(\mathbf{s}_i,\mathbf{s}_j)}{\max(m,n)}\right)
+
0.3 \frac{\mathrm{LCS}(\mathbf{s}_i,\mathbf{s}_j)}{\max(m,n)}
\right),
\end{equation}
where $\sigma(\cdot)$ denotes a sigmoid-based normalization function. The stroke-sequence similarity is then defined as
\begin{equation}
S_{\mathrm{stroke}}(c_i, c_j)
=
\max \left(
S_{\mathrm{edit}}(c_i, c_j),
S_{\mathrm{order}}(c_i, c_j)
\right).
\end{equation}

Finally, the overall character-level homoglyph similarity score is computed as
\begin{equation}
S_{\mathrm{char}}(c_i, c_j)
=
\max \left(
S_{\mathrm{vis}}(c_i, c_j),
S_{\mathrm{stroke}}(c_i, c_j)
\right).
\end{equation}
Equivalently, this aggregation corresponds to taking the maximum over perceptual hash similarity, SSIM similarity, stroke-sequence edit similarity, and DTW--LCS stroke-order similarity. We use the maximum operator because strong similarity under any one reliable view can already lead to practical confusion in financial document recognition, and high recall is preferred when constructing hard-negative candidates.

For candidate retrieval, we additionally use stroke-count similarity as an auxiliary cue for writing complexity. Let the stroke counts of $c_i$ and $c_j$ be $n_i$ and $n_j$, respectively:
\begin{equation}
S_{\mathrm{count}}(c_i, c_j)
=
1 -
\frac{|n_i - n_j|}{\max(n_i,n_j)}.
\end{equation}
This score is used to expand the candidate pool together with the stroke-order lexicon, especially for Chinese characters. However, it is not included in the final maximum aggregation, because similar writing complexity alone may introduce many visually unrelated characters.

Based on this mechanism, we first retrieve more than 200 candidate homoglyphs for each target character using the stroke-order and stroke-count lexicons. We then compute the final similarity score $S_{\mathrm{char}}(c_i,c_j)$ for each candidate pair and retain only those candidates whose score exceeds a fixed threshold of $0.6$. This threshold is used consistently in all experiments and empirically provides a practical balance between preserving visually confusable candidates and suppressing low-quality matches. In practice, after thresholding, each target character typically retains 5--30 high-confidence homoglyphs. This range is not manually imposed; rather, it emerges naturally from the score distribution after filtering.

The same overall retrieval-and-filtering framework is applied across Chinese characters, English letters, digits, and common symbols. However, the available similarity cues differ by character type. For Chinese characters, we use both rendered-glyph visual similarity and stroke-sequence similarity, with stroke-count similarity serving as an auxiliary retrieval cue. For English letters, digits, and symbols, rendered-glyph visual similarity serves as the primary criterion, since stroke-order and stroke-count information is either unavailable or less informative for these categories.

Representative examples of the resulting financial homoglyph vocabulary are provided in Appendix~\ref{appendix:financial_homoglyph_vocab}. The vocabulary covers common visual confusions among Chinese characters, English letters, digits, and symbols, and serves as the basis for constructing hard negative samples.

Given a document image $x$ and its correct annotation $y^{+}$, we generate hard negatives by replacing selected characters in $y^{+}$ with samples drawn from the corresponding homoglyph sets, for example replacing ``i'' with ``|''. This produces a hard negative text $y^{-}$. Unlike random perturbations, these negatives differ from the correct text at only a small number of visually confusable positions, and therefore better reflect the fine-grained recognition errors observed in financial documents.

This strategy offers two advantages. First, the resulting negative samples are domain-relevant and cover high-risk cases such as amount confusion, name ambiguity, and partial ID-number errors. Second, because the positive and negative texts remain lexically close, the constructed negatives are genuinely hard and thus impose stronger discriminative pressure during training. Using this procedure, we generate 10--30 negative samples for each positive example, yielding approximately 2 million triplets for contrastive learning.

\vspace{0.5em}
\noindent\textbf{Dual-Path Contrastive Objectives}
\vspace{0.5em}

Our task is ultimately a generation problem: recovering structured text from a document image. Under the VLM framework, this naturally involves at least two representation spaces that benefit from explicit supervision: the visual representation space produced by the image encoder, and the multimodal generative representation space formed after image--text fusion. The former affects perception of local glyphs, layouts, and structural cues, whereas the latter has a more direct influence on the quality of the generated output. Motivated by this observation, we define dual-path contrastive objectives.

For an input image $x_i$, we obtain patch-level visual representations from the vision encoder and apply average pooling to obtain a global image representation:
\begin{equation}
u_i = \mathrm{Pool}(f_{\mathrm{img}}(x_i)) \in \mathbb{R}^d.
\end{equation}

For the corresponding correct text $y_i^{+}$ and hard negative text $y_i^{-}$, we feed them into the language model, take the hidden states from the last layer, and apply average pooling:
\begin{equation}
v_i^{+} = \mathrm{Pool}(f_{\mathrm{text}}(y_i^{+})), \qquad
v_i^{-} = \mathrm{Pool}(f_{\mathrm{text}}(y_i^{-})).
\end{equation}

It is worth noting that $v_i^{+}$ and $v_i^{-}$ are used as label-side textual prototypes, rather than as direct simulations of the autoregressive decoding trajectory. The purpose of these prototypes is to provide contrastive supervision: image-conditioned representations are encouraged to align with the correct textual prototype and move away from homoglyph-corrupted prototypes. This design allows the contrastive objective to regularize the representation space without modifying the standard supervised fine-tuning path.

Beyond the visual branch, we further extract a multimodal generative representation by feeding the image and the target text into the VLM and pooling the last-layer hidden states:
\begin{equation}
g_i = \mathrm{Pool}(f_{\mathrm{vlm}}(x_i, y_i^{+})).
\end{equation}
Intuitively, $g_i$ summarizes the internal multimodal representation associated with generating the target output from the input image.

Before similarity computation, all representations are $\ell_2$-normalized. We then adopt a margin-based contrastive loss with batch-averaged negatives \citep{schroff2015facenet,khosla2020supervised}. For an anchor representation $z_i$, a positive representation $p_i$, and a set of negative representations $\{n_j\}_{j=1}^{N}$, we define a learnable scaling factor
\begin{equation}
s = \exp(\mathtt{logit\_scale}),
\end{equation}
and compute
\begin{equation}
\mathcal{L}_{\mathrm{triplet}}(\{z_i\}_{i=1}^N, \{p_i\}_{i=1}^N, \{n_j\}_{j=1}^N)
=
\frac{1}{N}
\sum_{i=1}^{N}
\max \left(
0,\,
s \cdot \frac{1}{N}\sum_{j=1}^{N} z_i^\top n_j
-
s \cdot z_i^\top p_i
+ m
\right),
\end{equation}
where $N$ is the batch size and $m$ is a margin hyperparameter, fixed to $0.2$ in our experiments.

Based on the dual-path design above, we define one loss on the visual branch and one on the multimodal generative branch:
\begin{equation}
\mathcal{L}_{\mathrm{triplet,img}} =
\mathcal{L}_{\mathrm{triplet}}(\{u_i\}_{i=1}^N, \{v_i^{+}\}_{i=1}^N, \{v_i^{-}\}_{i=1}^N),
\end{equation}
\begin{equation}
\mathcal{L}_{\mathrm{triplet,gen}} =
\mathcal{L}_{\mathrm{triplet}}(\{g_i\}_{i=1}^N, \{v_i^{+}\}_{i=1}^N, \{v_i^{-}\}_{i=1}^N).
\end{equation}

Here, $\mathcal{L}_{\mathrm{triplet,img}}$ directly regularizes the visual branch, encouraging the encoder to learn fine-grained representations better suited to financial documents. Meanwhile, $\mathcal{L}_{\mathrm{triplet,gen}}$ regularizes the multimodal generative space, improving the model's ability to suppress homoglyph errors during decoding. Together, these two objectives strengthen both visual discrimination and generation-time error resistance.

\vspace{0.5em}
\noindent\textbf{Joint Optimization with Supervised Fine-Tuning}
\vspace{0.5em}

Contrastive objectives alone are insufficient to preserve the model's sequence generation ability. We therefore jointly optimize the dual-path contrastive losses together with the standard supervised fine-tuning objective. For an input image $x$ and target sequence
\begin{equation}
y = (y_1, \dots, y_T),
\end{equation}
the SFT loss is
\begin{equation}
\mathcal{L}_{\mathrm{SFT}}
=
-
\sum_{t=1}^{T}
\log P(y_t \mid y_{<t}, x).
\end{equation}

The final objective is
\begin{equation}
\mathcal{L}_{\mathrm{total}}
=
\lambda_1 \mathcal{L}_{\mathrm{triplet,img}}
+
\lambda_2 \mathcal{L}_{\mathrm{triplet,gen}}
+
\lambda_3 \mathcal{L}_{\mathrm{SFT}},
\end{equation}
where $\lambda_1,\lambda_2,\lambda_3$ control the relative weights of the three terms.

\paragraph{Discussion.}
Unlike standard image--text contrastive learning, our method does not rely on massive weakly supervised image--text pairs. Instead, it models high-risk errors in financial document parsing through task-specific triplets of \emph{(image, correct text, hard negative text)}. Compared with random perturbation or generic negative sampling, hard negatives constructed via homoglyph substitution better reflect realistic recognition errors in financial deployments and therefore provide more informative training signals. Starting from Qwen3-VL pre-trained weights, this stage gradually adapts both the visual branch and the multimodal representation space while retaining the base model's general capabilities, thereby establishing a stronger representation foundation for subsequent reinforcement learning with domain-specific rewards.

\subsubsection{Reinforcement Learning Fine-Tuning with Domain-Specific Rewards}

After enhancing the model's visual and multimodal representations, we further optimize FinixDoc-VL end-to-end with reinforcement learning, shifting the objective from pure imitation of annotated outputs toward direct optimization of task-relevant utility. In financial document parsing, token-level cross-entropy alone has clear limitations. High-risk errors are not uniformly distributed across tokens; instead, they concentrate in critical elements such as amounts, dates, ID numbers, field names, table structures, and cross-region reading order. These factors are not adequately captured by generic text-similarity metrics, and a single supervised objective is insufficient to express the trade-offs among them.

We therefore adopt a GRPO-based reinforcement learning approach \citep{shao2024deepseekmath,guo2025deepseekr1} and design domain-specific reward functions tailored to financial document parsing. Furthermore, because text quality, detection quality, and reading order exhibit different reward characteristics and optimization dynamics, we employ a multi-stage RL strategy to improve training stability and overall alignment.

\begin{figure}[t]
    \centering
    \includegraphics[width=\textwidth]{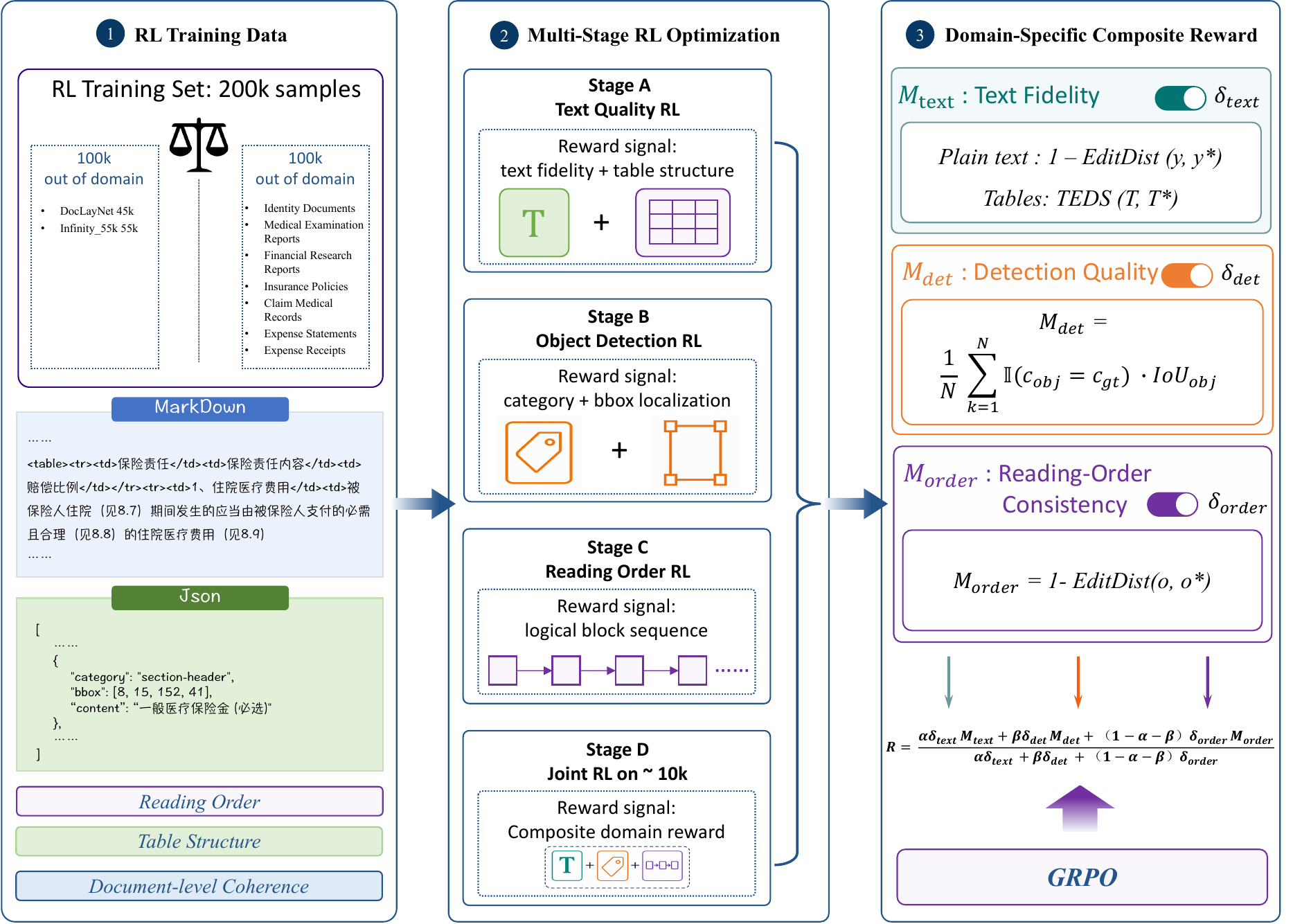}
    \caption{
    Overview of reinforcement learning fine-tuning with domain-specific rewards. The RL stage jointly optimizes JSON and Markdown generation through multi-stage sub-capability calibration followed by unified GRPO alignment. Task-dependent reward masks combine text fidelity, detection quality, and reading-order consistency according to the output format.
    }
    \label{fig:rl_overview}
\end{figure}

As summarized in Figure~\ref{fig:rl_overview}, the RL stage starts from dual task formulations over JSON and Markdown generation, proceeds through multi-stage sub-capability calibration, and finally performs unified GRPO alignment using task-dependent domain rewards.

\vspace{0.5em}
\noindent\textbf{Training Data and Task Formulation}
\vspace{0.5em}

Our RL training set contains 200k samples, including 100k out-of-domain samples and 100k in-domain financial samples. The out-of-domain portion is drawn primarily from public document parsing datasets and is used to preserve general layout understanding and open-domain structural generalization \citep{pfitzmann2022doclaynet,zhong2019publaynet,li2020docbank,wang2025infinityparser}. The in-domain portion covers common and high-value document types in real financial workflows and is used to strengthen performance on complex layouts, low-quality images, and high-risk fields. The detailed composition is shown in Table~\ref{tab:rl_data}.

\begin{table}[htbp]
\centering
\caption{Composition of the training data used in the reinforcement learning stage.}
\label{tab:rl_data}
\begin{tabular}{lll}
\toprule
Source & Dataset / Document Type & Size \\
\midrule
Out-of-domain & DocLayNet & 45k \\
Out-of-domain & Infinity-Doc-55K & 55k \\
In-domain & Identity Documents & 20k \\
In-domain & Medical Examination Reports & 10k \\
In-domain & Financial Research Reports & 10k \\
In-domain & Insurance Policies & 10k \\
In-domain & Claim Medical Records & 10k \\
In-domain & Expense Statements & 20k \\
In-domain & Expense Receipts & 20k \\
\midrule
\multicolumn{2}{l}{Total} & 200k \\
\bottomrule
\end{tabular}
\end{table}

During RL, we optimize two generation tasks jointly. The first is structured JSON generation. Given an input page, the model outputs a sequence of structural objects, each containing fields such as category, bounding box, and text content. This task evaluates layout detection, category prediction, localization, and consistency of structured outputs. The second is Markdown generation. Given an input page, the model generates the full page content in natural reading order, recovering paragraph structure, heading hierarchy, and table content as faithfully as possible, with tables represented in an HTML-compatible format. Compared with the JSON task, Markdown generation places greater emphasis on global textual coherence, reading-order reconstruction, and faithful recovery of complex tables.

These two tasks correspond to complementary usage patterns in practical systems. JSON output is well suited as an intermediate parsing layer for downstream rule engines, field extraction modules, and structured business systems. Markdown output is more suitable as a unified document representation for downstream LLM-based tasks such as question answering, summarization, and clause understanding. By optimizing both tasks, we encourage the model to acquire both fine-grained structural parsing ability and stronger document-level understanding.

\vspace{0.5em}
\noindent\textbf{Multi-Stage Optimization Strategy}
\vspace{0.5em}

We do not optimize all RL objectives jointly from the beginning. Instead, we adopt a multi-stage pipeline, which can be summarized as \emph{sub-capability calibration followed by unified alignment}. Specifically, we first optimize the three sub-objectives---text quality, object detection, and reading order---in separate stages, allowing the model to stabilize under more focused reward signals. We then perform joint RL on approximately 10k high-quality samples using a unified composite reward to achieve final alignment across capabilities.

This strategy is motivated by two considerations. First, heterogeneous rewards can interfere strongly during early RL training; for example, the model may sacrifice localization quality to increase text similarity, or degrade reading order to improve detection consistency. Multi-stage training helps reduce such instabilities. Second, the final joint stage explicitly teaches the model to balance multiple objectives, so that it can maintain text fidelity, structural integrity, localization quality, and reading-order consistency simultaneously at inference time. We have found this strategy to be particularly important for complex financial documents such as long tables, mixed-layout pages, and insurance clauses.

\vspace{0.5em}
\noindent\textbf{GRPO Optimization Objective}
\vspace{0.5em}

We adopt Group Relative Policy Optimization (GRPO) as the policy optimization algorithm \citep{shao2024deepseekmath,guo2025deepseekr1}. For an input sample $x$, the current policy $\pi_{\theta}$ generates a group of candidate outputs $\{y_i\}_{i=1}^{G}$, and each candidate is assigned a reward $R(x, y_i)$. We normalize the rewards within the group to obtain relative advantages:
\begin{equation}
A_i=\frac{R(x,y_i)-\mu_R}{\sigma_R+\varepsilon_R},
\end{equation}
where $\mu_R$ and $\sigma_R$ are the mean and standard deviation of the group rewards, respectively, and $\varepsilon_R$ is a small constant for numerical stability. The GRPO objective is
\begin{equation}
\mathcal{L}_{\mathrm{GRPO}}
=
-\mathbb{E}_{x,\{y_i\}}
\left[
\frac{1}{G}\sum_{i=1}^{G}
\min\left(
r_iA_i,\,
\mathrm{clip}(r_i,1-\epsilon_{\mathrm{clip}},1+\epsilon_{\mathrm{clip}})A_i
\right)
\right]
+\lambda_{\mathrm{KL}} D_{\mathrm{KL}}(\pi_{\theta} \parallel \pi_{\mathrm{ref}}),
\end{equation}
where
\begin{equation}
r_i=\frac{\pi_{\theta}(y_i|x)}{\pi_{\theta_{\mathrm{old}}}(y_i|x)}.
\end{equation}

Here, $\pi_{\mathrm{ref}}$ is a reference policy, $\epsilon_{\mathrm{clip}}$ is the clipping coefficient, and $\lambda_{\mathrm{KL}}$ controls the KL regularization strength. The KL term prevents the updated policy from drifting excessively during RL training. Compared with standard PPO formulations that often rely on explicit value estimation, GRPO constructs policy gradients directly from relative rewards within a sampled group. This is particularly suitable for structured document parsing, where the output space is complex and the reward naturally combines multiple heterogeneous metrics.

\needspace{6\baselineskip}
\vspace{0.5em}
\noindent\textbf{Domain-Specific Reward Design}
\vspace{0.5em}

To improve the practical utility of structured document parsing, we design a composite reward that jointly evaluates text fidelity, detection quality, and reading-order consistency. Because not every reward component applies to every output format, we introduce task-dependent binary masks $\delta_{\mathrm{text}}, \delta_{\mathrm{det}}, \delta_{\mathrm{order}} \in \{0,1\}$ and define
\begin{equation}
R
=
\frac{
\alpha \delta_{\mathrm{text}} M_{\mathrm{text}}
+
\beta \delta_{\mathrm{det}} M_{\mathrm{det}}
+
(1-\alpha-\beta)\delta_{\mathrm{order}} M_{\mathrm{order}}
}{
\alpha \delta_{\mathrm{text}}
+
\beta \delta_{\mathrm{det}}
+
(1-\alpha-\beta)\delta_{\mathrm{order}}
},
\end{equation}
where $M_{\mathrm{text}}$, $M_{\mathrm{det}}$, and $M_{\mathrm{order}}$ denote the rewards for text fidelity, detection quality, and reading-order consistency, respectively. The hyperparameters $\alpha,\beta \in [0,1]$ satisfy $\alpha+\beta \le 1$. For inactive reward terms, the corresponding mask is set to $0$, so unavailable components are excluded and the remaining weights are renormalized automatically. In practice, at least one reward term is active for every task, so the denominator is always non-zero. Table~\ref{tab:reward-hyperparams} summarizes the actual hyperparameter values and mask settings used in our training.

\begin{table}[!ht]
\centering
\caption{Reward hyperparameters and mask settings for the JSON and Markdown tasks.}
\label{tab:reward-hyperparams}
\begin{tabular}{lcccccc}
\toprule
Task & $\alpha$ & $\beta$ & $1-\alpha-\beta$ & $\delta_{\mathrm{text}}$ & $\delta_{\mathrm{det}}$ & $\delta_{\mathrm{order}}$ \\
\midrule
JSON     & 0.5 & 0.4 & 0.1 & 1 & 1 & 1 \\
Markdown & 0.5 & 0.4 & 0.1 & 1 & 0 & 1 \\
\bottomrule
\end{tabular}
\end{table}

The text-fidelity reward measures similarity between the prediction and the ground truth over both plain-text and table objects. For regular text objects, we use normalized edit distance \citep{levenshtein1966binary}:
\begin{equation}
M_{\mathrm{text,plain}} = 1 - \mathrm{EditDist}(y, y^{*}),
\end{equation}
where $y$ and $y^{*}$ are the predicted and reference text, respectively, and $\mathrm{EditDist}(\cdot,\cdot)\in[0,1]$ denotes normalized edit distance. For table objects, character-level similarity alone is insufficient because a table may contain mostly correct cell content while still breaking row--column relations or hierarchical structure. We therefore use Tree-Edit-Distance-based Similarity (TEDS) \citep{zhang1989simple,zhong2020image}:
\begin{equation}
M_{\mathrm{text,table}} = \mathrm{TEDS}(T, T^{*}),
\end{equation}
where $T$ and $T^{*}$ are the predicted and reference table structures, respectively. The overall text reward is then computed by weighted aggregation over all page objects:
\begin{equation}
M_{\mathrm{text}}
=
\frac{\sum_k w_k M_k}{\sum_k w_k},
\end{equation}
where $M_k$ is instantiated as either $M_{\mathrm{text,plain}}$ or $M_{\mathrm{text,table}}$ depending on the object type, and $w_k$ denotes the importance weight of the $k$-th object.

The detection reward is used primarily for JSON outputs, where the model must predict both structural categories and bounding boxes. Let prediction--ground-truth object pairs be matched by a predefined assignment rule.\footnote{In implementation, matching can be performed by maximum-IoU assignment or Hungarian matching, depending on the annotation format.} Let $\mathcal{P}$ denote the set of matched pairs, with $N_{\mathrm{match}}=|\mathcal{P}|$. For each matched pair $(\hat{o},o^{*})\in\mathcal{P}$, let $c_{\hat{o}}$ and $c_{o^{*}}$ be the predicted and ground-truth categories, and let $\mathrm{IoU}(\hat{o},o^{*})$ be the intersection-over-union of the corresponding boxes. We define
\begin{equation}
M_{\mathrm{det}}
=
\frac{1}{N_{\mathrm{match}}}
\sum_{(\hat{o},o^{*})\in\mathcal{P}}
\mathbb{I}\{c_{\hat{o}}=c_{o^{*}}\} \cdot \mathrm{IoU}(\hat{o},o^{*}),
\end{equation}
where $\mathbb{I}\{\cdot\}$ is the indicator function. This reward encourages the model to produce correct structural labels together with accurate localization.

The reading-order reward evaluates whether the model preserves the logical reading sequence of the document, which is particularly important for multi-column pages, mixed text--table layouts, and long documents. Let $o$ denote the predicted object or block sequence and $o^{*}$ the corresponding reference order. For Markdown outputs, this sequence is derived from the linearized block order in the rendered structure. We define
\begin{equation}
M_{\mathrm{order}} = 1 - \mathrm{EditDist}(o, o^{*}),
\end{equation}
where $\mathrm{EditDist}(\cdot,\cdot)$ is normalized to $[0,1]$. This reward explicitly constrains document-level ordering consistency and penalizes outputs whose local content is correct but whose global reading order is inconsistent.

\paragraph{Discussion.}
Our reward design is task-dependent. For JSON outputs, text fidelity, structural category, and localization are combined at the object level before page-level aggregation. For Markdown outputs, the reward emphasizes text fidelity, reading-order consistency, and structural preservation of tables through TEDS. Unlike RL approaches based on generic text-level metrics such as BLEU, ROUGE, or global edit distance, our method incorporates objectives that directly reflect the requirements of practical financial document parsing. In particular, the model is encouraged to produce accurate text, correct structural categories, precise bounding boxes, coherent reading order, and structurally faithful tables. This makes the RL stage a useful complement to SFT for improving structural consistency and practical usability.

%% file: content/3_dataset.tex
\section{Data Factory: A Human-in-the-Loop Production Pipeline for Financial Documents}

\subsection{Motivation}

High-quality structured data is fundamental to training and improving document parsing models. In the financial and insurance domain, however, data construction faces three major practical challenges.

First, \textbf{high domain specificity}: annotation requires accurate handling of financial terminology, clause hierarchies, and highly specialized document layouts.

Second, \textbf{large quality variation}: a single production pipeline must remain compatible with digitally native PDFs, scanned copies, and mobile-captured images of varying quality.

Third, \textbf{massive data scale}: historical business operations have accumulated corpora at the scale of millions of financial documents and tens of millions of document pages, while large-scale manual annotation remains prohibitively expensive.

To our knowledge, few existing industrial pipelines jointly address these three challenges in a unified design. Many focus on a single aspect of the problem, such as hard-example mining, multi-model agreement, or progressive annotation, without integrating model diversity, multi-stage refinement, and downstream quality feedback into a closed loop. To address this gap, we build an AI-driven, human-in-the-loop \textit{Data Factory} pipeline for producing structured training and evaluation data from large-scale raw financial documents. The system adopts a three-stage cascaded architecture that combines heterogeneous multi-model inference, model-based refinement, and rule-based validation, followed by confidence-aware routing for automatic acceptance, spot-checking, or mandatory expert review. In practice, this design shifts human effort from full manual annotation to targeted auditing and correction of high-risk samples, making large-scale data production more feasible.

\begin{figure}[t!]
  \centering
  \scalebox{1}[0.95]{%
    \includegraphics[trim=0cm 0cm 0cm 0cm, clip, width=\linewidth]{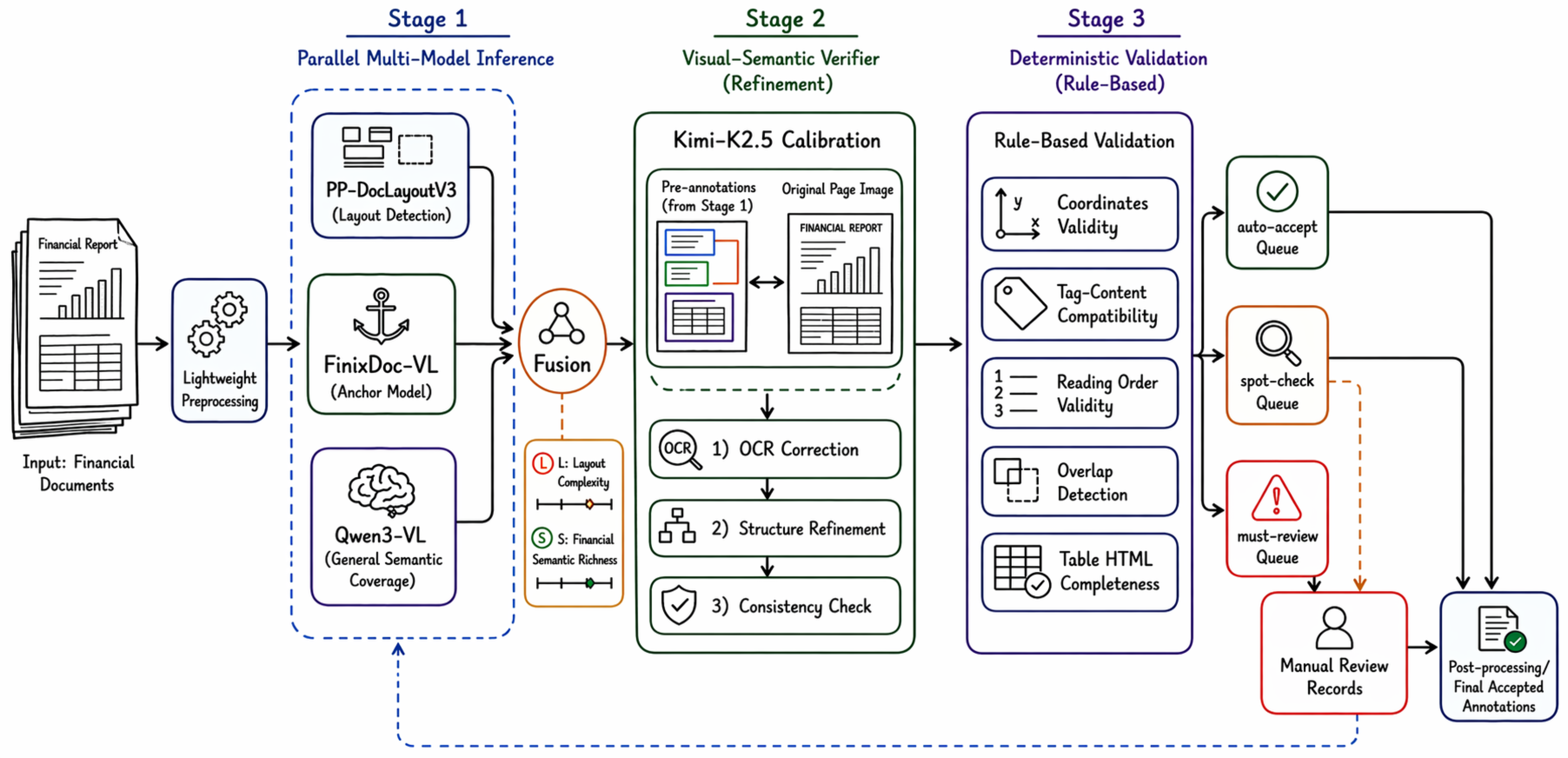}%
  }
  \caption{\textbf{Overview of the Data Factory pipeline.}}
  \label{fig:data-pipeline}
\end{figure}

\subsection{Pipeline Architecture and Core Mechanisms}

The Data Factory consists of lightweight preprocessing, a three-stage core pipeline, and post-processing. The main architecture and operating mechanisms are summarized below.

\vspace{0.5em}
\noindent\textbf{Stage 1: Heterogeneous Multi-Model Collaborative Inference and Multi-Dimensional Scoring}
\vspace{0.5em}

Each page first enters a parallel multi-model inference framework to extract both layout structures and semantic content. Within this framework, we utilize three distinct models: PP-DocLayoutV3, which is dedicated exclusively to stable layout detection without extracting textual content; FinixDoc-VL, which provides financial-domain parsing and schema-aligned structural priors; and Qwen3-VL-235B-A22B-Instruct, a large-scale general-purpose VLM that complements domain-specific priors with broader semantic coverage \citep{sun2025ppdoclayout,qwen2025qwen3vl}. Since FinixDoc-VL is the model most closely aligned with our target domain and annotation schema, its output is used as the anchor in the subsequent fusion stage. Specifically, objects proposed by the other models are matched to the FinixDoc-VL anchor output based on category compatibility, spatial overlap, and text similarity. Model-specific confidence and agreement signals are then aggregated to produce a unified candidate annotation for the current page.

Based on the comprehensive parsing results obtained from this multi-model fusion, the system evaluates each page by computing two page-level statistics: layout complexity $\mathcal{L}$ and financial semantic richness $\mathcal{S}$. Let the fused page-level annotation be denoted as
\begin{equation}
\mathcal{B}
=
\left\{
b_i =
(x_i^{(1)}, y_i^{(1)}, x_i^{(2)}, y_i^{(2)}, c_i, t_i)
\right\}_{i=1}^{n},
\end{equation}
where $b_i$ denotes the $i$-th fused layout region, $(x_i^{(1)}, y_i^{(1)}, x_i^{(2)}, y_i^{(2)})$ is its normalized bounding box, $c_i$ is its layout category, and $t_i$ is the extracted textual content. We further define the geometric region of $b_i$ as
\begin{equation}
r_i =
[x_i^{(1)}, x_i^{(2)}]
\times
[y_i^{(1)}, y_i^{(2)}],
\end{equation}
and its normalized area as
\begin{equation}
a_i =
|r_i|
=
(x_i^{(2)}-x_i^{(1)})(y_i^{(2)}-y_i^{(1)}).
\end{equation}

The layout complexity score $\mathcal{L}$ measures structural difficulty by considering layout region count, layout density, overlap complexity, and layout-type diversity. Let $s_{\mathrm{reg}}$, $s_{\mathrm{den}}$, $s_{\mathrm{ovl}}$, and $s_{\mathrm{div}}$ denote the corresponding normalized component scores. We compute $\mathcal{L}$ as
\begin{equation}
\mathcal{L}
=
0.30\,s_{\mathrm{reg}}
+
0.25\,s_{\mathrm{den}}
+
0.20\,s_{\mathrm{ovl}}
+
0.25\,s_{\mathrm{div}}.
\end{equation}

The normalized region count score is defined as
\begin{equation}
s_{\mathrm{reg}}
=
\min\left(
\frac{n}{N_{\max}},
1
\right),
\end{equation}
where $N_{\max}$ is the reference maximum number of layout regions per page.

The layout density score is defined as the union area covered by all fused layout regions:
\begin{equation}
s_{\mathrm{den}}
=
\left|
\bigcup_{i=1}^{n} r_i
\right|.
\end{equation}
Since all page coordinates are normalized to $[0,1]$, this value naturally lies within $[0,1]$.

The overlap complexity score measures the degree of spatial conflict among layout regions:
\begin{equation}
s_{\mathrm{ovl}}
=
\begin{cases}
0, & n < 2, \\[4pt]
\min\left(
\frac{
\sum_{i<j}|r_i \cap r_j|
}{
\tau_{\mathrm{o}} \sum_{i=1}^{n} a_i + \epsilon
},
1
\right), & n \geq 2,
\end{cases}
\end{equation}
where $\tau_{\mathrm{o}}$ is a normalization constant and $\epsilon$ is a small constant used to avoid division by zero.

The layout-type diversity score is defined as the fraction of predefined layout categories that appear on the page:
\begin{equation}
s_{\mathrm{div}}
=
\frac{
\left|
\{c_i \mid b_i \in \mathcal{B}\}
\right|
}{
|\mathcal{C}|
},
\end{equation}
where $\mathcal{C}$ denotes the predefined set of layout categories.

Meanwhile, the financial semantic richness score $\mathcal{S}$ measures semantic informativeness based on the extracted text, including OCR text volume, financial term density, and content diversity. Let
\begin{equation}
\mathcal{T}
=
\{b_i \in \mathcal{B} \mid |t_i| > 0\}
\end{equation}
be the set of regions with non-empty textual content, let $\mathcal{F}$ denote a predefined financial-domain term dictionary, and let $\mathcal{C}_{\mathrm{text}} \subseteq \mathcal{C}$ denote the subset of layout categories that can contain textual information.

Let $s_{\mathrm{text}}$, $s_{\mathrm{fin}}$, and $s_{\mathrm{cont}}$ denote the normalized OCR text volume, financial-term density, and content diversity scores, respectively. The semantic richness score is computed as
\begin{equation}
\mathcal{S}
=
0.35\,s_{\mathrm{text}}
+
0.40\,s_{\mathrm{fin}}
+
0.25\,s_{\mathrm{cont}}.
\end{equation}

The OCR text volume score is normalized by a reference maximum text length $T_{\max}$:
\begin{equation}
s_{\mathrm{text}}
=
\min\left(
\frac{
\sum_{b_i \in \mathcal{T}} |t_i|
}{
T_{\max}
},
1
\right).
\end{equation}

The financial-term density score is defined as the normalized proportion of financial-domain terms in the extracted text:
\begin{equation}
s_{\mathrm{fin}}
=
\min\left(
\frac{1}{\rho_{\mathrm{f}}}
\cdot
\frac{
\sum_{w \in \operatorname{tokens}(\mathcal{T})}
\mathbb{I}\{w \in \mathcal{F}\}
}{
|\operatorname{tokens}(\mathcal{T})| + \epsilon
},
1
\right),
\end{equation}
where $\rho_{\mathrm{f}}$ is the reference financial-term density. In our implementation, we set $\rho_{\mathrm{f}}=0.10$, meaning that a page whose financial-domain terms account for 10\% of all extracted tokens reaches the maximum normalized financial-term density.

The content diversity score is defined as the fraction of text-bearing layout categories present on the page:
\begin{equation}
s_{\mathrm{cont}}
=
\frac{
\left|
\{c_i \mid b_i \in \mathcal{T}\}
\right|
}{
|\mathcal{C}_{\mathrm{text}}|
}.
\end{equation}

Finally, these two heuristic composite scores are used for post-hoc data selection and stratified sampling. The overall page value score, denoted by $V_{\mathrm{page}}$, is defined as
\begin{equation}
V_{\mathrm{page}}
=
0.5\,\mathcal{L}
+
0.5\,\mathcal{S}.
\end{equation}
Pages are assigned to four sampling quadrants using thresholds $\theta_{\mathcal{L}}$ and $\theta_{\mathcal{S}}$. A page is regarded as high-complexity if $\mathcal{L} \geq \theta_{\mathcal{L}}$ and high-semantics if $\mathcal{S} \geq \theta_{\mathcal{S}}$. These two binary decisions form the four-quadrant sampling space used in the subsequent data selection stage.

\vspace{0.5em}
\noindent\textbf{Stage 2: Cascaded Large-Model Calibration and Refinement}
\vspace{0.5em}

After the multi-model fusion stage produces the initial candidate annotations, these outputs are not directly treated as final labels. Instead, the fused annotations are passed, together with the original page image and the corresponding extracted textual evidence, into a cascaded refinement pipeline driven by a more powerful multimodal foundation model, Kimi-K2.5 \citep{kimiteam2026k25}. In this stage, Kimi-K2.5 serves as a visual-semantic verifier and refiner that re-examines the pre-annotated results against the original document image, the recognized text, and the page-level structural context. This design allows the system to correct residual errors introduced during heterogeneous model collaboration, such as cross-model disagreement, noisy object proposals, inaccurate bounding boxes, OCR inconsistencies, duplicated regions, and hallucinated layout elements.

The refinement process is organized as a three-step cascade, where each step focuses on a specific class of annotation errors while preserving the outputs of previous stages for traceability:
\begin{itemize}
    \item \textbf{Step 1 (OCR):} performs low-level textual verification and correction. The model compares the pre-annotated textual content with the visual evidence in the original page, revising OCR errors, recovering missing or truncated text, and correcting basic category misclassifications when the textual pattern clearly conflicts with the assigned label.
    \item \textbf{Step 2 (Struct):} focuses on structural calibration of the annotation layout. Kimi-K2.5 verifies whether each bounding box accurately covers the intended visual region, adjusts misaligned or fragmented boxes, removes duplicated proposals, and optimizes the reading order according to the spatial organization of the page. For complex table pages, this step activates a table integrity protection mechanism to avoid breaking coherent table structures during box adjustment and order refinement.
    \item \textbf{Step 3 (Consist):} conducts final global consistency checking across the refined annotations. The model inspects whether the page-level annotation set is coherent with the original image and with the outputs accumulated from earlier stages, mitigating common VLM failure modes such as hallucinated layout elements, unsupported semantic assignments, missing key regions, and unresolved cross-model inconsistencies.
\end{itemize}

By introducing this larger-model calibration stage after the initial heterogeneous fusion, the pipeline separates broad candidate generation from high-precision annotation correction. The first stage maximizes recall by aggregating complementary model outputs, while the second stage uses Kimi-K2.5 to improve precision and structural reliability through image-grounded refinement. All intermediate annotations and revision records are preserved, enabling end-to-end traceability and providing explicit evidence for the subsequent validation and quality routing stage.

\vspace{0.5em}
\noindent\textbf{Stage 3: Structured Validation and Quality Feedback Loop}
\vspace{0.5em}

After model refinement, the annotations enter a programmatic validation stage driven by deterministic rules. The system performs structured checks and lightweight rule-based repairs along multiple dimensions, including coordinate validity, tag-content compatibility, reading-order consistency, box-overlap anomalies, and table HTML completeness. Based on the validation results, the system aggregates multiple confidence signals, such as inter-model agreement, refinement stability across stages, OCR consistency, and rule-violation counts, and routes each page into one of three queues: \texttt{auto-accept}, \texttt{spot-check}, and \texttt{must-review}. Pages routed to \texttt{spot-check} are manually audited by sampling, while pages routed to \texttt{must-review} undergo mandatory human review and correction before final acceptance.

The resulting human inspection records, including corrected annotations, error categories, and reviewer comments, are then fed back into the Stage 1 pre-annotation process. In subsequent production rounds, these feedback signals are used to update annotation guidelines, refine model prompts, adjust fusion heuristics, and calibrate confidence thresholds, thereby improving the precision of initial candidate annotations. This feedback mechanism allows human review experience to continuously improve upstream pre-annotation quality, while the confidence-aware routing mechanism concentrates costly human inspection on high-risk samples where model uncertainty remains high.

\subsection{Annotation Standards and Domain-Specific Ambiguity Resolution}

Reliable large-scale data production requires a stable annotation standard. Our annotation guidelines were jointly developed by domain experts and the annotation team and were designed to remain compatible with the open-source MinerU 2.5 framework \citep{niu2025mineru25}. The entire pipeline uses 10 standard page-element tags: \texttt{page-header}, \texttt{page-footer}, \texttt{title}, \texttt{section-header}, \texttt{text}, \texttt{table}, \texttt{picture}, \texttt{caption}, \texttt{footnote}, and \texttt{other}.

To better match the characteristics of financial documents, we do not treat \texttt{list-item} and \texttt{formula} as standalone classes. In particular, standalone mathematical expressions occur only sporadically across financial documents, accounting for a negligible fraction of all layout elements; treating them as a dedicated category therefore introduces schema overhead without commensurate benefit. Accordingly, list items are categorized semantically as either \texttt{text} or \texttt{section-header}, while mathematical expressions are serialized as inline \LaTeX{} spans within the \texttt{text} category rather than being annotated as a separate layout class. This design reduces schema fragmentation and better matches the dominant usage patterns in financial documents. In addition, the Data Factory includes dedicated rules for several high-frequency ambiguity cases commonly observed in financial documents:

\begin{itemize}
    \item \textbf{Boundary between hierarchical headings (\texttt{section-header}) and body text (\texttt{text}):} if a text span begins with hierarchical numbering (e.g., ``I.'', ``1.1'', ``(1)'') and functions as a discourse-level section introducer in context, it is classified as \texttt{section-header}. By contrast, the detailed provisions listed under such numbering are classified as \texttt{text}.
    \item \textbf{Distinguishing figure/table descriptions (\texttt{caption}) from footnotes (\texttt{footnote}):} a \texttt{caption} must be adjacent to the figure or table that it describes, whereas text placed at the bottom of the page and serving as a page-level disclaimer or supplementary note is classified as \texttt{footnote}.
    \item \textbf{Table integrity protection (\texttt{table}):} for complex tables with many merged cells, such as insurance premium rate tables and cash value tables, the system prevents a single logical table from being split into multiple independent \texttt{text} regions. Instead, a dedicated table-preservation mechanism is applied, and the table is serialized uniformly as HTML \texttt{<table>} content that preserves merged-cell structure.
\end{itemize}

For geometric representation, the entire pipeline uses normalized coordinates $[x_1, y_1, x_2, y_2] \in [0,1]^4$, with denormalization performed only at final export. For all elements except \texttt{picture}, the \texttt{content} field retains the associated text, structural markers, or serialized content (e.g., Markdown heading markers such as \#, \#\#, or HTML table content).

\subsection{Data Sampling Strategy and Final Output}

The Data Factory is initialized with a pool of approximately 10 million raw financial document pages. After the three-stage pipeline and confidence-aware routing, pages are retained in the validated page pool only if they are either automatically accepted with sufficient confidence or accepted after the applicable spot-check or mandatory human-review procedure. Instead of applying random sampling to this validated pool, we perform \textbf{four-quadrant stratified sampling} based on the layout complexity $\mathcal{L}$ and financial semantic richness $\mathcal{S}$ computed in Stage 1:

\begin{itemize}
    \item \textbf{Q1 (High Complexity + High Semantics):} fully retained as the core backbone set for downstream training, especially reinforcement learning.
    \item \textbf{Q2 (Low Complexity + High Semantics):} up-sampled to increase coverage of long-tail financial semantics.
    \item \textbf{Q3 (High Complexity + Low Semantics):} not directly used in standard training; instead, reserved as a candidate pool for robustness-oriented training, such as error-correction training and hard-case augmentation.
    \item \textbf{Q4 (Low Complexity + Low Semantics):} heavily down-sampled or filtered out.
\end{itemize}

Through this quality-aware data selection strategy, we ultimately retain approximately 100,000 pages of validated in-domain training data, corresponding to an overall retention rate of roughly 1\% at the page level. This curated distribution helps reduce overfitting to a narrow set of document layout patterns during training.

The final in-domain dataset covers major high-frequency scenarios in financial workflows, including Identity Documents, Medical Examination Reports, Financial Research Reports, Insurance Policies, Claim Medical Records, Expense Statements, and Expense Receipts. In the subsequent reinforcement learning (RL) stage, these 100,000 pages of refined in-domain data are mixed with an additional approximately 100,000 pages of out-of-domain public data, such as DocLayNet and Infinity-Doc-55K, resulting in a final training set of roughly 200,000 pages \citep{pfitzmann2022doclaynet,wang2025infinityparser}. This data mixture improves both cross-domain generalization and in-domain robustness for financial document parsing.

%% file: content/4_evaluation.tex
\section{Evaluation}

\subsection{Evaluation Benchmarks}

We evaluate our core model, FinixDoc-VL, against representative document parsing baselines on two benchmarks: OmniDocBench v1.5 and our proposed FinixDocBench. For the ultra-large-page subset of FinixDocBench, we further evaluate the complete FinixDoc system, including the Document Preprocessing Tool Layer and its split-then-merge (tiling-and-merging) module. Together, these evaluations cover a broad spectrum of settings, ranging from general document parsing to highly specialized and deployment-critical financial scenarios.

\subsubsection{OmniDocBench v1.5}

OmniDocBench v1.5 is a public benchmark for general document parsing \citep{ouyang2025omnidocbench}. It contains 1,355 pages spanning diverse document types, including academic papers, technical reports, and commercial contracts, with an average of over 1,100 tokens per page. The benchmark covers common layout patterns such as multi-column text, complex tables, mathematical formulas, and nested document structures. It primarily evaluates foundational parsing capabilities under relatively clean, digitally native conditions.

The official overall score is defined as:
\begin{equation}
S_{\mathrm{ODB}}
=
\frac{
100(1 - E_{\mathrm{text}})
+
\mathrm{TEDS}_{\mathrm{table}}
+
\mathrm{CDM}_{\mathrm{formula}}
}{3},
\end{equation}
where $E_{\mathrm{text}}$, $\mathrm{TEDS}_{\mathrm{table}}$, and $\mathrm{CDM}_{\mathrm{formula}}$ denote the text edit score, the table TEDS score, and the formula CDM score, respectively.

Although FinixDoc-VL is not explicitly optimized for mathematical formula parsing, we still evaluate the formula task on OmniDocBench v1.5 to provide a complete and fair comparison. By contrast, some compared baselines report formula results while others do not; therefore, we use their official overall scores where available. To enable a more comparable evaluation on the shared dimensions, we further introduce an auxiliary metric that excludes the formula component, denoted as $S_{\mathrm{ODB}}^{\mathrm{wo\mbox{-}formula}}$:
\begin{equation}
S_{\mathrm{ODB}}^{\mathrm{wo\mbox{-}formula}}
=
\frac{
100(1 - E_{\mathrm{text}})
+
\mathrm{TEDS}_{\mathrm{table}}
}{2}.
\end{equation}

In the following analysis on OmniDocBench v1.5, we use $S_{\mathrm{ODB}}^{\mathrm{wo\mbox{-}formula}}$ as the primary comparison metric. This choice is also consistent with the industrial focus of our work: mathematical formulas account for only a negligible fraction of real-world financial documents and rarely constitute the practical bottleneck in downstream deployment.

\subsubsection{FinixDocBench}

As discussed in the Introduction, the main challenges in real-world financial document parsing arise from two sources: low-quality camera-captured inputs and complex large-scale pages, such as ultra-long documents and dense large tables. Existing public benchmarks provide only limited coverage of these challenging settings, leaving a non-trivial gap between benchmark performance and real deployment effectiveness.

To bridge this gap, we construct \textbf{FinixDocBench}, a benchmark specifically designed for financial document parsing. It systematically evaluates model capabilities on digitally native financial documents, camera-captured documents, and ultra-large pages. Compared with existing public benchmarks, FinixDocBench has three distinguishing characteristics:
\begin{enumerate}
    \item \textbf{Comprehensive quality coverage:} it includes both digitally native documents and authentically noisy camera-captured images;
    \item \textbf{Broad scale variation:} it spans standard pages, dense large tables, and ultra-long page-like documents;
    \item \textbf{High-frequency financial scenarios:} it emphasizes common real-world use cases such as insurance policies, reimbursement materials, and financial research reports.
\end{enumerate}

FinixDocBench supports two complementary evaluation paradigms:
\begin{itemize}
    \item \textbf{Structured parsing evaluation:} based on JSON annotations, focusing on layout structure, table structure, and reading order consistency;
    \item \textbf{Full-text parsing evaluation:} based on full-page Markdown representations, focusing on text fidelity and overall readability.
\end{itemize}

The annotation format is fully consistent with our training data and follows a unified 10-class tag system. Each bounding box annotation contains normalized coordinates $[x_1, y_1, x_2, y_2] \in [0,1]^4$, a category label, and the corresponding transcribed text content. Importantly, all evaluation samples in FinixDocBench are constructed separately from the training corpus used by Data Factory and are not used in contrastive adaptation, supervised fine-tuning, or reinforcement learning.

\subsubsection{Data Composition}

The full FinixDocBench dataset contains approximately 5,000 pages, covering the primary document types and quality-degradation patterns encountered in real financial workflows. It is organized into four evaluation tracks: \textbf{FinixDigital}, \textbf{FinixPhoto}, \textbf{FinixHuge}, and \textbf{FinixInner}. The FinixDigital, FinixPhoto, and FinixHuge tracks are constructed around public-facing document scenarios, while FinixInner covers high-frequency internal financial workflow scenarios. To support broader research, we release a compliance-reviewed subset of FinixDocBench alongside this technical report.

The overall composition is summarized in Table~\ref{tab:finixdocbench_composition}.

\begin{table}[htbp]
\centering
\caption{\textbf{Data composition of FinixDocBench.}}
\label{tab:finixdocbench_composition}
\resizebox{\linewidth}{!}{
\begin{tabular}{lllc}
\toprule
\textbf{Document Type} & \textbf{Source Type} & \textbf{Evaluation Track} & \textbf{Volume (Pages)} \\
\midrule
Financial Reports \& Insurance Clauses & Digitally Native & FinixDigital & 500 \\
Medical Receipts & Mobile Captured & FinixPhoto & 300 \\
Large-Scale Financial Pages & Mixed Quality & FinixHuge & 200 \\
Identity Documents & Mobile Captured & FinixInner & 800 \\
Medical Examination Reports & Mobile Captured & FinixInner & 800 \\
Claim Medical Records & Mobile Captured & FinixInner & 800 \\
Expense Statements & Mobile Captured & FinixInner & 800 \\
Expense Receipts & Mobile Captured & FinixInner & 800 \\
\bottomrule
\end{tabular}
}
\end{table}

FinixDocBench is organized into four complementary evaluation tracks:
\begin{itemize}
    \item \textbf{FinixDigital:} a track for digitally native financial documents, covering financial research reports and insurance clauses (approximately 500 pages). The samples are high-definition PDFs from public-facing financial document scenarios. Characterized by dense semantic content, specialized terminology, and deep hierarchical structures, this track evaluates the model's ability to recognize domain terms, parse financial tables, and reconstruct document hierarchies accurately.

    \item \textbf{FinixPhoto:} a track for camera-captured documents, containing approximately 300 pages of receipt images from the CHIP 2022 public-scenario source, all re-annotated in our unified format. Unlike the original competition annotations, our benchmark uses newly constructed full-page annotations and full-text ground truth under a unified financial document parsing schema. This subset introduces realistic low-quality factors, including motion blur, perspective distortion, and printing artifacts, and primarily evaluates model robustness under authentic mobile-capture conditions. Although prior exposure by some external models to the original CHIP 2022 source cannot be fully ruled out, these evaluation samples are not used in our own training pipeline.

    \item \textbf{FinixHuge:} a track for ultra-large page documents, containing approximately 200 extremely high-resolution pages with dense and complex layouts, such as ultra-long insurance clauses and very large dense tables from public-facing document scenarios. This subset primarily evaluates parsing stability and structural preservation under multi-region inference. \textit{This subset is evaluated using the complete FinixDoc system, including the Document Preprocessing Tool Layer's split-then-merge module, rather than the standalone FinixDoc-VL model.}

    \item \textbf{FinixInner:} a track for high-frequency financial and insurance workflow documents, containing approximately 4,000 camera-captured pages. It covers identity documents, medical examination reports, claim medical records, expense statements, and expense receipts. These samples exhibit overlapping degradations such as shadow occlusion, blur, glare, and perspective distortion. FinixInner is used only for held-out evaluation and is never included in model training.
\end{itemize}

Across all FinixDocBench tracks, annotations are first produced as pre-annotations by our AI-driven Data Factory pipeline, and are then verified and corrected through rigorous manual quality assurance before being finalized for evaluation.





\subsubsection{Evaluation Metrics}

FinixDocBench deliberately excludes a standalone formula parsing metric. In real financial scenarios, formulas are sparse and rarely constitute the deployment bottleneck. Instead, the practical determinants of usability are text fidelity, table structure recovery, reading order consistency, and robustness to low-quality inputs.

Accordingly, we define the composite overall score of FinixDocBench as:
\begin{equation}
S_{\mathrm{FDB}}
=
\frac{
100(1 - E_{\mathrm{text}})
+
\mathrm{TEDS}_{\mathrm{table}}
+
100(1 - E_{\mathrm{order}})
}{3},
\end{equation}
where:
\begin{itemize}
    \item \textbf{Text Edit Distance ($\downarrow$):} $E_{\mathrm{text}}$, the character-level normalized edit distance computed on the full-text Markdown output \citep{levenshtein1966binary};
    \item \textbf{Table TEDS / Table TEDS-S ($\uparrow$):} $\mathrm{TEDS}_{\mathrm{table}}$ and $\mathrm{TEDS\mbox{-}S}_{\mathrm{table}}$, Tree-Edit-Distance-based Similarity scores computed on table structure predictions, where TEDS-S is invariant to the spatial order of cells \citep{zhang1989simple,zhong2020image};
    \item \textbf{Read Order Edit Distance ($\downarrow$):} $E_{\mathrm{order}}$, the normalized edit distance computed on the serialized sequence of structural blocks, measuring consistency between predicted and reference reading orders \citep{levenshtein1966binary}.
\end{itemize}

In the overall score, we use $\mathrm{TEDS}_{\mathrm{table}}$ as the table metric, while $\mathrm{TEDS\mbox{-}S}_{\mathrm{table}}$ is reported as an auxiliary metric for order-invariant table evaluation.

\subsection{Main Results}

We first validate the effectiveness of our visual adaptation strategy through an isolated ablation study. We then present comprehensive evaluations on both general and domain-specific benchmarks to demonstrate the generality, robustness, and deployment relevance of FinixDoc-VL.

\subsubsection{Ablation on ViT Contrastive Learning: Driving License Recognition}

To validate the effectiveness of the visual adaptation strategy introduced in Section~\ref{sec:contrastive_learning}, we conduct a targeted ablation study on an internal driving license recognition task. This task requires extracting key fields from suboptimal camera-captured photographs and therefore serves as a representative Low-Quality Zone setting that directly stresses the robustness of the visual encoder. The purpose of this experiment is to isolate the contribution of visual adaptation under realistic low-quality capture conditions, rather than to benchmark full financial document parsing performance.

All evaluated models are fine-tuned on approximately 20k task-specific samples and tested on a private set of roughly 4k samples. The train and test sets are fully disjoint. The evaluation metric is \textit{Document-level Accuracy}, under which a sample is counted as correct only if every required field is extracted without error. Since this experiment is designed to isolate the effect of the visual adaptation stage, its results are reported separately and are not intended for direct comparison with the benchmark results in the subsequent sections.

\begin{table}[htbp]
  \centering
  \caption{\textbf{Ablation results on the internal driving license recognition task.} Performance is measured by Document-level Accuracy.}
  \label{tab:vit_ablation}
  \begin{tabular}{lc}
    \toprule
    \textbf{Model} & \textbf{Document-level Accuracy ($\uparrow$)} \\
    \midrule
    Donut & 0.922 \\
    MinerU & 0.924 \\
    Qwen3-VL-2B & 0.917 \\
    Qwen3-VL-4B & 0.929 \\
    \textbf{Qwen3-VL-4B (ViT-only adapted)} & \textbf{0.935} \\
    \bottomrule
  \end{tabular}
\end{table}

As shown in Table~\ref{tab:vit_ablation}, adapting only the visual encoder already yields a clear and consistent improvement in authentic camera-captured scenarios. Although the raw base-model ranking is not strictly monotonic with parameter scale on this task, the adapted 4B model achieves the best result overall, indicating that domain-aligned visual representation learning is more important than model size alone under this type of low-quality input. This result provides direct evidence that stronger domain-adapted visual representations are a necessary foundation for the downstream improvements in structural parsing and semantic reconstruction achieved by FinixDoc-VL.

\subsubsection{General Document Parsing: OmniDocBench v1.5}

We first evaluate the foundational parsing capability of FinixDoc-VL on OmniDocBench v1.5 \citep{ouyang2025omnidocbench}. Table~\ref{tab:omnidocbench} compares FinixDoc-VL against both specialized document parsing models \citep{cui2026paddleocrvl15,niu2025mineru25,wei2026deepseekocr2,dotsocr2025,duan2026glmocr,fireredteam2026fireredocr,youtu2026parsing} and large-scale general-purpose VLMs \citep{qwen2025qwen3vl,qwen2026qwen35blog,kimiteam2026k25}. For compared baselines, we report official benchmark numbers where available; $S_{\mathrm{ODB}}^{\mathrm{wo\mbox{-}formula}}$ is recomputed from the reported text and table metrics under the same formula for shared-dimension comparison.

\begin{table}[htbp]
\centering
\caption{\textbf{Performance on OmniDocBench v1.5.}}
\resizebox{\linewidth}{!}{
\begin{tabular}{lccccccc}
\toprule
\textbf{Model} &
\textbf{$S_{\mathrm{ODB}}^{\mathrm{wo\mbox{-}formula}}$} $\uparrow$ &
\textbf{$S_{\mathrm{ODB}}$} $\uparrow$ &
\textbf{$E_{\mathrm{text}}$} $\downarrow$ &
\textbf{$\mathrm{CDM}_{\mathrm{formula}}$} $\uparrow$ &
\textbf{$\mathrm{TEDS}_{\mathrm{table}}$} $\uparrow$ &
\textbf{$\mathrm{TEDS\mbox{-}S}_{\mathrm{table}}$} $\uparrow$ &
\textbf{$E_{\mathrm{order}}$} $\downarrow$ \\
\midrule
Qwen3-VL-4B & 86.90 & 86.78 & 0.055 & 86.55 & 79.29 & 84.26 & 0.084 \\
\textbf{FinixDoc-VL} & 92.65 & 87.40 & 0.042 & 76.89 & 89.50 & 92.76 & 0.055 \\
\midrule
\rowcolor[gray]{0.95}
\multicolumn{8}{l}{\textit{Specialized Models}} \\
DeepSeek-OCR-2 & 90.35 & 89.17 & 0.049 & 86.85 & 85.60 & 90.06 & 0.060 \\
Dots.OCR & 90.99 & 88.41 & 0.048 & 83.22 & 86.78 & 90.62 & 0.053 \\
MinerU 2.5 & 91.97 & 90.93 & 0.045 & 88.86 & 88.44 & 92.42 & 0.044 \\
FireRed-OCR & 92.61 & 92.07 & \textbf{0.035} & 90.98 & 88.72 & 92.38 & 0.041 \\
Youtu-Parsing & 94.45 & 93.37 & 0.042 & 91.22 & 93.10 & 96.47 & \textbf{0.026} \\
GLM-OCR & \textbf{94.70} & 94.35 & 0.045 & 93.65 & \textbf{93.89} & \textbf{96.50} & 0.047 \\
PaddleOCR-VL-1.5 & 94.63 & \textbf{94.50} & \textbf{0.035} & \textbf{94.21} & 92.76 & 95.79 & 0.042 \\
\midrule
\rowcolor[gray]{0.95}
\multicolumn{8}{l}{\textit{General-purpose VLMs}} \\
Qwen3.5-397B-A17B & 88.72 & 88.34 & 0.055 & 87.60 & 82.93 & 88.70 & 0.080 \\
Qwen3-VL-235B-A22B-Instruct & 89.66 & 89.15 & 0.069 & 88.14 & 86.21 & 90.55 & 0.068 \\
Kimi-K2.5 & 90.54 & 89.33 & 0.065 & 86.92 & 87.57 & 91.82 & 0.084 \\
\bottomrule
\end{tabular}
}
\vspace{0.5em}
\label{tab:omnidocbench}
\end{table}

FinixDoc-VL demonstrates strong foundational capability on this general benchmark, achieving $S_{\mathrm{ODB}}^{\mathrm{wo\mbox{-}formula}}=92.65$. In particular, it achieves $\mathrm{TEDS\mbox{-}S}_{\mathrm{table}}=92.76$, together with $E_{\mathrm{order}}=0.055$, indicating accurate table structure recovery and strong reading-order consistency on clean, digitally native documents.

Specialized document models such as PaddleOCR-VL-1.5 and GLM-OCR remain ahead on this benchmark. This outcome is unsurprising: OmniDocBench is concentrated in the Benchmark-Converged Zone of our capability matrix, where documents are relatively clean and the task emphasis is strongly aligned with mature layout-analysis and formula-aware pipelines. Nevertheless, FinixDoc-VL substantially outperforms its initialization base Qwen3-VL-4B across all shared non-formula metrics, confirming that our training recipe preserves general parsing capability while improving the model's structural and textual fidelity.

By contrast, very large general-purpose VLMs remain noticeably weaker on structural document parsing, especially in reading order and table reconstruction. The purpose of this evaluation is therefore not to claim absolute SOTA on a general clean benchmark, but to establish that FinixDoc-VL maintains a solid general parsing foundation before moving into the more challenging financial scenarios.

\subsubsection{Financial Digitally Native Documents: FinixDigital}

\begin{table}[htbp]
\centering
\caption{\textbf{Performance on FinixDigital.}}
\label{tab:finixdigital}
\resizebox{\linewidth}{!}{
\begin{tabular}{lccccc}
\toprule
\textbf{Model} &
\textbf{$S_{\mathrm{FDB}}$} $\uparrow$ &
\textbf{$E_{\mathrm{text}}$} $\downarrow$ &
\textbf{$\mathrm{TEDS}_{\mathrm{table}}$} $\uparrow$ &
\textbf{$\mathrm{TEDS\mbox{-}S}_{\mathrm{table}}$} $\uparrow$ &
\textbf{$E_{\mathrm{order}}$} $\downarrow$ \\
\midrule
Qwen3-VL-4B & 80.18 & 0.145 & 76.04 & 81.23 & 0.210 \\
\textbf{FinixDoc-VL} & \textbf{93.19} & \textbf{0.039} & \textbf{92.07} & \textbf{93.67} & 0.086 \\
\midrule
\rowcolor[gray]{0.95}
\multicolumn{6}{l}{\textit{Specialized Models}} \\
DeepSeek-OCR-2 & 82.80 & 0.139 & 90.00 & 92.32 & 0.277 \\
FireRed-OCR & 83.10 & 0.119 & 87.10 & 89.14 & 0.259 \\
PaddleOCR-VL-1.5 & 85.41 & 0.116 & 86.12 & 88.26 & 0.183 \\
GLM-OCR & 86.28 & 0.121 & 89.44 & 90.99 & 0.185 \\
Youtu-Parsing & 89.26 & 0.091 & 87.79 & 90.85 & 0.109 \\
Dots.OCR & 90.36 & 0.058 & 89.78 & 92.26 & 0.129 \\
MinerU 2.5 & 92.96 & 0.045 & 91.18 & 92.70 & \textbf{0.078} \\
\midrule
\rowcolor[gray]{0.95}
\multicolumn{6}{l}{\textit{General-purpose VLMs}} \\
Qwen3.5-397B-A17B & 84.90 & 0.119 & 87.30 & 89.51 & 0.207 \\
Kimi-K2.5 & 85.05 & 0.119 & 85.95 & 88.24 & 0.189 \\
Qwen3-VL-235B-A22B-Instruct & 87.26 & 0.076 & 82.77 & 85.44 & 0.134 \\
\bottomrule
\end{tabular}
}
\end{table}

As defined by the Document Parsing Capability Matrix, digitally native financial documents belong largely to the Benchmark-Converged Zone. Even in this relatively mature setting, FinixDoc-VL achieves the best $S_{\mathrm{FDB}}$ score of 93.19. This advantage is driven by highly accurate character-level recognition on dense financial terminology ($E_{\mathrm{text}}=0.039$) together with strong structural fidelity in complex financial tables ($\mathrm{TEDS}_{\mathrm{table}}=92.07$).

A more important observation is that strong performance on general clean benchmarks does not directly transfer to the financial domain. For example, GLM-OCR and PaddleOCR-VL-1.5 perform extremely well on OmniDocBench, but their scores on FinixDigital drop to 86.28 and 85.41, respectively. This gap indicates that financial documents impose domain-specific requirements beyond generic layout parsing, including specialized terminology, hierarchical clauses, and complex table semantics. FinixDoc-VL closes this gap effectively by combining general VLM robustness with targeted financial-domain adaptation.

\subsubsection{Camera-Captured Documents: FinixPhoto}

\begin{table}[htbp]
\centering
\caption{\textbf{Performance on FinixPhoto.}}
\label{tab:finixphoto}
\resizebox{\linewidth}{!}{
\begin{tabular}{lccccc}
\toprule
\textbf{Model} &
\textbf{$S_{\mathrm{FDB}}$} $\uparrow$ &
\textbf{$E_{\mathrm{text}}$} $\downarrow$ &
\textbf{$\mathrm{TEDS}_{\mathrm{table}}$} $\uparrow$ &
\textbf{$\mathrm{TEDS\mbox{-}S}_{\mathrm{table}}$} $\uparrow$ &
\textbf{$E_{\mathrm{order}}$} $\downarrow$ \\
\midrule
Qwen3-VL-4B & 54.28 & 0.408 & 50.13 & 63.04 & 0.465 \\
\textbf{FinixDoc-VL} & \textbf{67.03} & \textbf{0.276} & 69.08 & \textbf{77.69} & 0.404 \\
\midrule
\rowcolor[gray]{0.95}
\multicolumn{6}{l}{\textit{Specialized Models}} \\
PaddleOCR-VL-1.5 & 41.28 & 0.512 & 34.54 & 46.72 & 0.595 \\
MinerU 2.5 & 43.08 & 0.513 & 35.54 & 48.58 & 0.550 \\
DeepSeek-OCR-2 & 43.20 & 0.459 & 30.69 & 43.52 & 0.552 \\
GLM-OCR & 45.82 & 0.521 & 50.47 & 60.22 & 0.609 \\
FireRed-OCR & 47.20 & 0.487 & 38.50 & 53.72 & 0.482 \\
Dots.OCR & 52.57 & 0.399 & 44.90 & 56.92 & 0.473 \\
Youtu-Parsing & 60.90 & 0.345 & 59.01 & 66.23 & 0.418 \\
\midrule
\rowcolor[gray]{0.95}
\multicolumn{6}{l}{\textit{General-purpose VLMs}} \\
Qwen3.5-397B-A17B & 62.58 & 0.384 & 62.04 & 71.82 & 0.359 \\
Qwen3-VL-235B-A22B-Instruct & 62.65 & 0.359 & 63.55 & 72.13 & \textbf{0.397} \\
Kimi-K2.5 & 65.55 & 0.325 & \textbf{70.16} & 77.34 & 0.410 \\
\bottomrule
\end{tabular}
}
\end{table}

FinixPhoto corresponds to the \textbf{Low-Quality Zone} in our capability matrix, where documents are affected by blur, perspective distortion, compression artifacts, and other authentic capture noise. As shown in Table~\ref{tab:finixphoto}, the performance landscape changes substantially relative to digitally native benchmarks.

First, specialized document models degrade sharply. Systems that perform strongly on clean digital benchmarks drop into the 40--50 range under mobile-capture conditions, indicating limited robustness once the input distribution departs from the clean document assumptions under which such models are typically trained.

Second, large general-purpose VLMs become considerably more competitive in this setting. Benefiting from broader and more diverse pretraining, models such as Kimi-K2.5 and Qwen3-VL-235B-A22B-Instruct retain substantially stronger robustness to visual corruption than most specialized document parsers.

Third, FinixDoc-VL achieves the best $S_{\mathrm{FDB}}$ score of 67.03 and the best $\mathrm{TEDS\mbox{-}S}_{\mathrm{table}}$ score of 77.69. Compared with its base model Qwen3-VL-4B, it improves the $S_{\mathrm{FDB}}$ score by 12.75 points, demonstrating that our financial-domain visual adaptation and business-aligned RL optimization effectively convert the inherent robustness of a general VLM into stronger domain-specific parsing performance. This result is especially important because it validates the central motivation of this work: in real financial deployment, robustness on low-quality documents is often more consequential than marginal gains on already converged clean benchmarks.

\subsubsection{Ultra-Large Document Pages: FinixHuge}

FinixHuge targets the \textbf{Underexplored Large-Scale Zone} in our Document Parsing Capability Matrix, where the dominant challenge is no longer conventional OCR difficulty alone, but the joint failure of perception, generation, and integration under extreme page scale. In this setting, the primary practical bottleneck is often not only how accurately a model parses a page, but whether the page can be processed into a valid structured output at all. Since the key issue is full-pipeline processability rather than standard single-pass page parsing alone, we adopt a usability-oriented metric suite tailored to oversized-page scenarios.

Accordingly, this subset evaluates \textbf{end-to-end system usability} rather than raw model capability under strictly controlled input settings. We compare the complete \textbf{FinixDoc} system against two representative baselines, \textbf{Qwen3-VL-235B-A22B-Instruct} and \textbf{GLM-OCR}, both of which perform direct single-pass inference on the original input image \citep{qwen2025qwen3vl,duan2026glmocr}. By contrast, FinixDoc employs the \textbf{split-then-merge} strategy described in Section~\ref{sec:overall_architecture} and processes oversized images through a lightweight three-stage pipeline of \textit{Dynamic Routing $\rightarrow$ Structure-Aware Segmentation $\rightarrow$ Reconstruction}. Specifically, inputs exceeding the effective VLM token budget are routed to the segmentation branch, partitioned into multiple structure-aware sub-regions for parallel parsing, and then merged into a page-level result through direction-aware integration followed by a final integrity check.

In addition to \textbf{$E_{\mathrm{text}}$ ($\downarrow$)}, \textbf{$\mathrm{TEDS}_{\mathrm{table}}$ / $\mathrm{TEDS\mbox{-}S}_{\mathrm{table}}$ ($\uparrow$)}, and \textbf{$E_{\mathrm{order}}$ ($\downarrow$)}, we further report \textbf{Success Rate}. A page is counted as successful if the method returns a syntactically valid, non-empty page-level parsing result without runtime failure, severe truncation, or format errors that prevent downstream evaluation. Quality metrics are computed over successfully parsed pages, while failed pages are reflected by the Success Rate. This separation allows us to distinguish parsing quality from basic processability, which is critical in the ultra-large-page regime.

\begin{table}[htbp]
\centering
\caption{\textbf{Performance on FinixHuge.} FinixDoc denotes the complete
system, including the Document Preprocessing Tool Layer's split-then-merge
(tiling-and-merging) module.}
\label{tab:finixhuge}
\resizebox{\linewidth}{!}{
\begin{tabular}{lccccccc}
\toprule
\textbf{Model} &
\textbf{Success Rate} $\uparrow$ &
\textbf{$S_{\mathrm{FDB}}$} $\uparrow$ &
\textbf{$E_{\mathrm{text}}$} $\downarrow$ &
\textbf{$\mathrm{TEDS}_{\mathrm{table}}$} $\uparrow$ &
\textbf{$\mathrm{TEDS\mbox{-}S}_{\mathrm{table}}$} $\uparrow$ &
\textbf{$E_{\mathrm{order}}$} $\downarrow$ \\
\midrule
\textbf{FinixDoc} & \textbf{0.92} & \textbf{68.23} & \textbf{0.357} & 57.09 & 60.10 & \textbf{0.167} \\
Qwen3-VL-235B-A22B-Instruct & 0.68 & 34.85 & 0.847 & 47.05 & \textbf{63.20} & 0.578 \\
GLM-OCR & 0.34 & 38.06 & 0.816 & \textbf{59.39} & 62.43 & 0.636 \\
\bottomrule
\end{tabular}
}
\end{table}

As shown in Table~\ref{tab:finixhuge}, FinixDoc achieves a Success Rate of \textbf{0.92}, successfully parsing the vast majority of pages in this subset, while Qwen3-VL-235B-A22B-Instruct and GLM-OCR reach \textbf{0.68} and \textbf{0.34}, respectively. This gap indicates that, in the oversized-page regime, direct single-pass inference frequently fails to produce a valid parse result. In such cases, processability becomes a central bottleneck before fine-grained recognition quality can even be assessed.

On the metrics most directly tied to end-to-end usability, FinixDoc achieves the best $S_{\mathrm{FDB}}$ score of \textbf{68.23}, the lowest $E_{\mathrm{text}}$ of \textbf{0.357}, and the lowest $E_{\mathrm{order}}$ of \textbf{0.167}. These results suggest that the perception--generation--integration chain remains sufficiently coherent after partitioning and reconstruction, preserving strong textual fidelity and reading-order consistency.

On the two table-level metrics, $\mathrm{TEDS}_{\mathrm{table}}$ and $\mathrm{TEDS\mbox{-}S}_{\mathrm{table}}$, FinixDoc is competitive with the baselines but does not lead. This near-parity is itself notable, because the comparison is shaped by two confounding factors that systematically disadvantage FinixDoc. First, the baseline scores are computed only over the subset of pages that can be successfully processed by single-pass inference, which introduces a survivorship bias toward easier inputs. In contrast, FinixDoc produces valid outputs for a much larger and more difficult portion of the benchmark. Second, the split-then-merge pipeline tends to produce longer and more structurally complete table predictions. As a result, roughly \textbf{25\%} of FinixDoc's table cases exceed the TEDS evaluator's \textbf{6-minute timeout threshold}; under the evaluation protocol, such timed-out cases are assigned a tree-edit distance of \textbf{1.0}, corresponding to a TEDS score of \textbf{0}. This mechanically lowers the averaged table metrics and reflects evaluator-side computational limits on long, complex predictions rather than a direct reconstruction-quality deficit.

Taken together, these results support the view that ultra-large document parsing is fundamentally a system-level problem. Simply scaling model size is insufficient once page resolution and output length exceed the effective operating range of single-pass VLM inference. The \textit{Dynamic Routing $\rightarrow$ Structure-Aware Segmentation $\rightarrow$ Reconstruction} pipeline enables FinixDoc to substantially improve processability in the Underexplored Large-Scale Zone of our capability matrix.

\subsubsection{Internal Financial Domain Evaluation: FinixInner}
\label{sec:finixinner}

\textbf{FinixInner} is an internal held-out evaluation track containing 4,000 camera-captured documents from high-frequency financial and insurance workflows. Following the same evaluation protocol as FinixDocBench, we report the \textbf{Overall} score for each document category, scaled to the range $[0,100]$ for consistency with the preceding benchmark results.

\begin{table}[htbp]
\centering
\caption{\textbf{Performance on FinixInner.}}
\label{tab:finixinner}
\resizebox{\linewidth}{!}{
\begin{tabular}{lcccccc}
\toprule
\textbf{Model} & \textbf{Overall} & \textbf{Claim Medical Records} & \textbf{Expense Statements} & \textbf{Expense Receipts} & \textbf{Medical Examination Reports} & \textbf{Identity Documents} \\
\midrule
Qwen3-VL-4B & 66.42 & 50.24 & 64.85 & 70.55 & 60.21 & 86.26 \\
\textbf{FinixDoc-VL} & \textbf{84.08} & \textbf{85.42} & \textbf{80.79} & 85.15 & \textbf{78.38} & 90.64 \\
\midrule
\rowcolor[gray]{0.95}
\multicolumn{7}{l}{\textit{Specialized Models}} \\
DeepSeek-OCR-2 & 52.77 & 42.40 & 54.37 & 66.92 & 53.64 & 46.51 \\
FireRed-OCR & 64.93 & 50.23 & 62.67 & 66.24 & 69.45 & 76.08 \\
PaddleOCR-VL-1.5 & 61.06 & 45.17 & 65.19 & 62.86 & 56.30 & 75.79 \\
MinerU 2.5 & 62.53 & 55.85 & 62.76 & 65.28 & 73.35 & 55.41 \\
GLM-OCR & 70.26 & 62.45 & 62.68 & 58.25 & 76.43 & 91.51 \\
Dots.OCR & 72.54 & 64.36 & 71.44 & 73.16 & 72.99 & 80.77 \\
Youtu-Parsing & 78.73 & 62.81 & 74.55 & \textbf{88.08} & 75.31 & 92.90 \\
\midrule
\rowcolor[gray]{0.95}
\multicolumn{7}{l}{\textit{General-purpose VLMs}} \\
Qwen3.5-397B-A17B & 70.83 & 58.24 & 73.61 & 75.33 & 61.22 & 85.77 \\
Kimi-K2.5 & 76.91 & 76.32 & 73.35 & 80.75 & 66.71 & 87.41 \\
Qwen3-VL-235B-A22B-Instruct & 76.32 & 60.08 & 79.27 & 80.86 & 68.30 & \textbf{93.10} \\
\bottomrule
\end{tabular}
}
\end{table}

Table~\ref{tab:finixinner} shows that FinixDoc-VL delivers the strongest and most balanced performance across the FinixInner benchmark. In particular, it achieves the best results on the three most challenging subsets: \textit{Claim Medical Records} (85.42), \textit{Expense Statements} (80.79), and \textit{Medical Examination Reports} (78.38). These categories are characterized by dense domain terminology, complex table structures, non-trivial reading orders, and severe quality degradation caused by mobile capture. The strong gains on these subsets indicate that FinixDoc-VL is especially effective in scenarios where visual robustness and structural consistency must be maintained simultaneously.

Several observations are noteworthy. First, specialized document models exhibit substantial variance across document categories. While some remain highly competitive on standardized formats---for example, Youtu-Parsing achieves the best result on \textit{Expense Receipts} (88.08) and also performs strongly on \textit{Identity Documents} (92.90)---their performance deteriorates noticeably on more heterogeneous and structurally complex categories such as \textit{Claim Medical Records}. This pattern suggests that strong results on standardized or cleaner layouts do not necessarily transfer to real-world financial documents with compounded visual and structural difficulties.

Second, very large general-purpose VLMs are competitive on template-driven categories. In particular, Qwen3-VL-235B-A22B-Instruct achieves the best score on \textit{Identity Documents} (93.10), likely benefiting from stronger general visual robustness and broader template priors acquired during large-scale pretraining. However, its score on \textit{Claim Medical Records} drops to 60.08, indicating that general-purpose capability alone is insufficient for reliably handling dense, noisy, and structurally demanding financial documents.

Overall, the FinixInner results reinforce our central claim: progress in industrial financial document parsing should not be judged solely by performance on clean or widely used benchmarks. Instead, it should be measured by whether a system can robustly process low-quality, high-complexity documents encountered in real deployment environments. Under this criterion, FinixDoc-VL demonstrates the strongest overall robustness and the best practical readiness among the evaluated models.

%% file: content/5_related_work.tex
\section{Related Work}

\subsection{Traditional OCR Methods and Document Parsing Pipelines}

Traditional document parsing systems typically adopt a modular pipeline architecture, decomposing the overall task into multiple stages such as optical character recognition (OCR), layout analysis, table structure parsing, key information extraction, and rule-based post-processing \citep{smith2007tesseract,zhang2024documentparsingunveiled,paddleocr2025technical}. These methods have high engineering maturity and are well suited to scenarios with stable templates, regular layouts, and clearly defined task boundaries. As a result, they have long formed the backbone of industrial document digitization systems, including invoice processing, form understanding, and document archiving. OCR systems such as PaddleOCR and the toolchains built around them remain core components of many production systems today \citep{paddleocr2025technical}.

However, pipeline-based methods also suffer from several well-known limitations. First, different sub-modules usually require separate design, training, and hyperparameter tuning, resulting in substantial system integration complexity and high maintenance costs. Second, errors from upstream OCR or layout detection can propagate and amplify in later stages such as structure recovery and field extraction, thereby constraining the end-to-end performance ceiling. Third, these methods often rely heavily on domain-specific rules, template assumptions, and human heuristics, which limits their transferability across templates and application scenarios. With the development of end-to-end document understanding methods, an increasing number of studies have begun to reduce their dependence on explicit OCR pipelines, instead reformulating document parsing as direct generation from page images to structured outputs \citep{kim2022donut,lee2023pix2struct,blecher2023nougat}. Representative works such as Donut demonstrated that OCR-free end-to-end document understanding is feasible, laying the groundwork for subsequent unified generative document parsing methods \citep{kim2022donut}.

\subsection{General Vision-Language Models for Document Understanding}

In recent years, the rapid development of general vision-language models (VLMs) has significantly advanced document understanding \citep{bai2025qwen25vl,qwen2025qwen3vl,glm2025vl}. Unlike traditional pipelines that separately address OCR, structure recovery, and semantic extraction, general VLMs tend to jointly model visual appearance, textual content, spatial layout, and cross-modal semantic relationships within a unified framework. This enables them to support a wide range of tasks, including page-level understanding, document question answering, structured extraction, and content generation. Representative model families include Qwen-VL and Kimi-VL, which have shown strong generalization and cross-task transfer ability across various document understanding tasks \citep{bai2025qwen25vl,qwen2025qwen3vl,kimiteam2025kimivl}.

Compared with traditional OCR pipelines, a key advantage of general VLMs is their unified modeling capability: they can directly generate structured outputs from raw document images, simplifying system design and providing a more direct interface for downstream reasoning tasks \citep{kim2022donut,lee2023pix2struct,bai2025qwen25vl}. This is particularly useful in scenarios where parsed document content must be passed to large language models (LLMs) for question answering, summarization, or decision support. More importantly, due to their larger scale, broader pretraining coverage, and stronger open-world visual priors, general VLMs often exhibit better robustness than document-specific models under distribution shifts such as blur, shadows, perspective distortion, and other image degradations. However, because they are not explicitly optimized for document parsing, they may still exhibit insufficient accuracy on dense tables, complex layouts, fine-grained character recognition, and long structured outputs. When visual evidence is insufficient, they may also rely excessively on language priors and generate plausible but ungrounded outputs, leading to hallucinations \citep{li2023evaluatingpope,liu2024hallucinationsurvey}. In high-stakes financial document parsing scenarios, such behavior is often unacceptable.

\subsection{Specialized Models for Document Parsing and OCR}

In addition to general VLMs, a growing number of models specifically optimized for document parsing and OCR have emerged \citep{cui2026paddleocrvl15,niu2025mineru25,wei2026deepseekocr2,dotsocr2025,duan2026glmocr}. These methods are typically designed around document-specific challenges, including dense text recognition, layout structure recovery, reading order reconstruction, table parsing, and structured result generation. Representative models include MinerU 2.5, Dots.OCR, GLM-OCR, DeepSeek-OCR-2, and PaddleOCR-VL-1.5. Compared with general multimodal models, these specialized systems usually incorporate stronger document priors and place greater emphasis on structural fidelity, page-level reconstruction quality, and parsing stability.

On mainstream benchmarks, specialized document models usually perform strongly on high-quality scans, digitally native documents, and relatively simple page layouts \citep{ouyang2025omnidocbench,cui2026paddleocrvl15,niu2025mineru25,dotsocr2025,duan2026glmocr}. They often surpass general-purpose models in tasks such as table parsing, reading order recovery, and preserving structural consistency. However, these gains are largely concentrated on clean and regular document distributions. Many existing methods are trained and evaluated mainly on digitally native pages, scanned copies, or synthetic data, while camera-captured documents---common in real-world business workflows---remain underrepresented. As a result, even though specialized models have achieved impressive benchmark performance, their robustness, stability, and scalability in real-world financial settings, especially in low-quality and large-scale scenarios, still require further improvement.

Overall, existing work has made substantial progress on clean and moderately complex document parsing, but the capability boundary under low-quality, large-scale, and risk-sensitive financial scenarios remains insufficiently explored. This gap motivates both the capability matrix and the system design proposed in this work.

%% file: content/6_conclusion.tex
\section{Conclusion}

In this report, we study the key challenges of document parsing in real-world financial scenarios. We analyze why current methods can achieve strong benchmark performance yet still fall short in practical deployment, especially when documents are low-quality, structurally complex, or extremely large. To make this gap explicit from an industrial perspective, we introduce a \textit{Document Parsing Capability Matrix}. To address it in practice, we present \textbf{FinixDoc}, an end-to-end agentic document parsing system tailored to the requirements of financial applications, together with a domain-adapted training recipe, an AI-driven human-in-the-loop Data Factory pipeline, and a benchmark designed for realistic financial deployment settings.

We argue that progress in financial document parsing should not be measured solely by performance on conventional benchmarks concentrated in the Benchmark-Converged Zone. Instead, evaluation should prioritize stability, robustness, and usability when models face low-quality, high-complexity, and large-scale documents. From this perspective, FinixDoc not only provides a practical parsing system for financial applications, but also highlights the need to rethink how document parsing systems are evaluated and improved in industrial settings.

Several promising directions remain for future work. First, in the Low-Quality and Ambiguous-Unrecoverable Zones, developing reliable generation strategies based on uncertainty estimation and refusal mechanisms will be crucial for financial scenarios, where missing a field is often preferable to extracting an incorrect one. Second, progress in the Underexplored Large-Scale Zone will likely require a more unified multimodal architecture that can jointly model fine-grained local perception, global document structure, and consistency across distant regions. Finally, we will continue expanding FinixDocBench to cover a broader range of real-world financial sub-scenarios and more fine-grained task settings, so as to encourage more systematic benchmarking of industrial document parsing capabilities.

Taken together, the capability matrix, the FinixDoc system, the associated training recipe, the human-in-the-loop Data Factory pipeline, and the evaluation benchmarks introduced in this work provide a practical foundation for advancing financial document parsing under real deployment constraints. More broadly, we hope this work can help shift document parsing research beyond already converged clean benchmarks and toward more robust, verifiable, and deployment-oriented capabilities.

%% file: content/appendix.tex
\section{Implementation Details of the Split-then-Merge Strategy}
\label{appendix:tiling_merging}

As discussed in Section~\ref{sec:overall_architecture}, the \textbf{Document Preprocessing Tool Layer} in FinixDoc includes a lightweight \textbf{split-then-merge} strategy for handling the \textit{Underexplored Large-Scale Zone} of the Document Parsing Capability Matrix, namely ultra-large-page documents that are difficult to process reliably through a single native VLM call. This strategy is not intended to replace the core parsing capability of FinixDoc-VL. Instead, it serves as an external engineering supplement that improves the tractability and stability of oversized-image parsing in practical deployment.

The design objective of this strategy is to transform ultra-large-page parsing from a single monolithic inference step into a modular pipeline that supports \textbf{adaptive routing}, \textbf{local-region parsing}, and \textbf{page-level result integration}. In practice, this design allows the system to preserve the native direct-parsing path for standard pages, while activating a structured split-then-merge process only when page scale or layout complexity makes single-pass inference unreliable.

\begin{figure}[htbp]
  \centering
  \includegraphics[width=\linewidth]{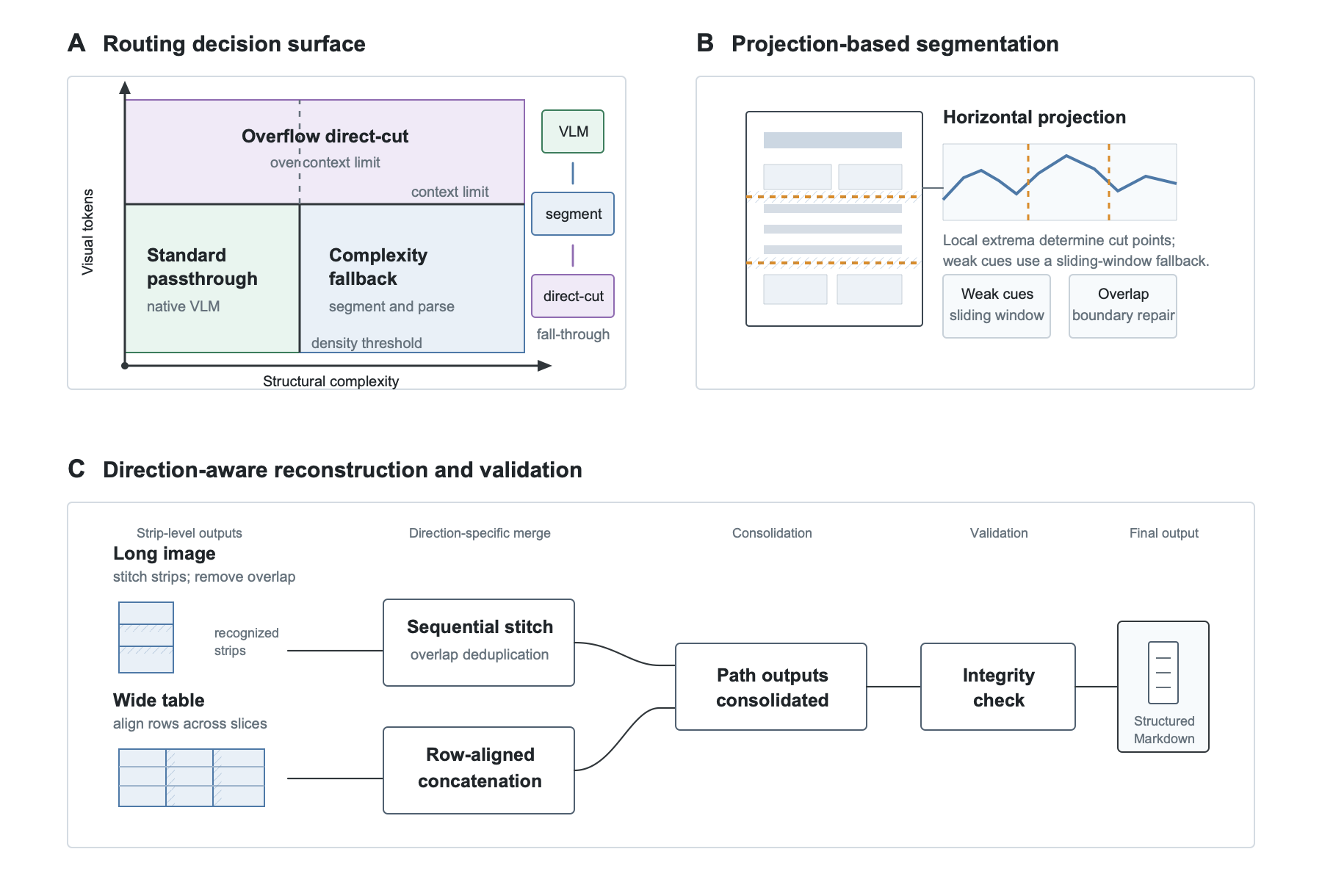}
  \caption{\textbf{Internal pipeline of the split-then-merge strategy in the Document Preprocessing Tool Layer.} The strategy follows a three-stage process: Dynamic Routing, Structure-Aware Segmentation, and Reconstruction. Depending on token feasibility and structural complexity, an input page is processed either by direct VLM parsing or by a split-then-merge path for ultra-large-page handling.}
  \label{fig:huge_matrix}
\end{figure}

As illustrated in Figure~\ref{fig:huge_matrix}, the strategy follows a three-stage internal pipeline:
\textit{Dynamic Routing $\rightarrow$ Structure-Aware Segmentation $\rightarrow$ Reconstruction}. Oversized images are first analyzed by a lightweight routing stage, then either sent directly to FinixDoc-VL or partitioned into multiple local regions for independent parsing, and finally merged into a unified page-level output.

\subsection{Dynamic Routing for Ultra-Large Pages}

Given an input image, the strategy first extracts a set of coarse page-level features, including pixel scale, aspect ratio, and estimated information density. Based on these signals, the page is dispatched in the two-dimensional space of \textit{Pixel Scale} $\times$ \textit{Structural Complexity} through a three-way routing strategy:

\begin{itemize}
    \item \textbf{Overflow-to-Splitting.} If the estimated visual token count exceeds the effective input budget of the underlying VLM, the page is routed directly to the split branch.

    \item \textbf{Direct Parsing.} If the token count remains within the feasible range and the page layout is relatively regular, the image is processed through the native FinixDoc-VL path and the result is returned directly.

    \item \textbf{Complexity Fallback.} If the token count is nominally acceptable but the page exhibits excessive semantic density or structural complexity---for example, dense large tables, ultra-long clauses, or mixed text-table layouts---the strategy falls back to the split branch to avoid unstable single-pass parsing.
\end{itemize}

In our implementation, the token-load estimate is derived from the effective visual patch count under the native visual tokenizer, while the structural complexity score is computed using lightweight heuristics based on page geometry, projection statistics, and coarse content-density indicators. This routing strategy preserves efficiency on ordinary pages while providing a fail-safe path for difficult oversized inputs. Rather than forcing all pages through segmented inference, the system activates splitting only when necessary.

\subsection{Structure-Aware Splitting Strategy}

For pages routed to the split branch, the strategy does not simply apply uniform equal-size partitioning. Instead, it uses a \textbf{structure-aware segmentation strategy} based on binarized orthogonal projection, with the goal of choosing cut locations that are better aligned with document layout.

This strategy includes three key mechanisms:

\begin{itemize}
    \item \textbf{Structure-Aligned Cutting.} According to the geometric profile of the page, the strategy computes horizontal or vertical projection maps and selects cut points near local extrema. After binarization, foreground density is accumulated along the target axis to form the projection profile used for cut-point selection. In practice, these locations tend to correspond to whitespace bands or low-density regions, which reduces the probability of splitting text lines, clause boundaries, or table rows across split regions.

    \item \textbf{Weak-Structure Fallback.} When projection signals are weak or ambiguous, the strategy falls back to a sliding-window splitting scheme. In this mode, cut positions are still chosen using projection-aware heuristics, so that the method remains applicable even on pages without strong regular layout cues.

    \item \textbf{Split Overlap.} Adjacent split regions preserve a fixed overlap region. This overlap provides local boundary context for subsequent deduplication, boundary repair, and page-level integration, especially when long text spans or table structures cross split boundaries.
\end{itemize}

Through this design, ultra-large pages are decomposed into a set of local regions that better preserve reading continuity and structural integrity than naive uniform splitting, while remaining suitable for independent parsing by FinixDoc-VL and amenable to parallel execution in deployment.

\subsection{Reconstruction and Merging}

After all local regions are parsed independently, the strategy performs page-level reconstruction through \textbf{direction-aware merging}. The merging strategy is selected according to the dominant geometric orientation of the page:

\begin{itemize}
    \item \textbf{Vertical merging for ultra-long pages.} For long-form pages, local outputs are stitched sequentially along the vertical direction. Overlap-aware deduplication is used to suppress repeated fragments and repair boundary inconsistencies introduced during splitting.

    \item \textbf{Horizontal merging for ultra-wide layouts.} For extremely wide pages, especially dense large tables, local outputs are integrated by row-aware horizontal concatenation, using overlap regions and line-level alignment cues to preserve row correspondence as much as possible and improve cell continuity across adjacent regions.
\end{itemize}

The merged result is then passed through a final integrity-check stage, which verifies global consistency, suppresses duplicated content, repairs minor structural discontinuities, and ensures that the final page-level output conforms to the target Markdown representation used throughout FinixDoc. This stage is particularly important for preserving content completeness and reading-order consistency on FinixHuge-style pages.

\subsection{End-to-End Control Flow}

For clarity, we summarize the full inference logic of the split-then-merge strategy in Algorithm~\ref{alg:tiling_merging_pipeline}.

\begin{algorithm}[htbp]
\caption{Split-then-Merge Pipeline for Ultra-Large Pages}
\label{alg:tiling_merging_pipeline}
\small
\begin{algorithmic}[1]
\Require Input image $I$, base parser $\mathcal{M}$, token budget $\tau$
\Ensure Structured Markdown output $Y$

\State Extract page-level features $f(I)$, including pixel scale, aspect ratio, and information density
\State Estimate effective token load $\hat{t}(I)$ and structural complexity score $\hat{c}(I)$

\If{$\hat{t}(I) > \tau$}
    \State Route $I$ to the split branch
\ElsIf{$\hat{t}(I) \le \tau$ \textbf{and} $\hat{c}(I)$ is low}
    \State $Y \gets \mathcal{M}(I)$ \Comment{Direct FinixDoc-VL parsing}
    \State \Return $Y$
\Else
    \State Route $I$ to the split branch \Comment{Complexity fallback}
\EndIf

\State Determine the primary splitting direction based on page geometry
\State Compute orthogonal projection profiles along the target axis
\State Select structure-aware cut points; if unstable, switch to sliding-window fallback
\State Generate overlapping local regions $\{I_k\}_{k=1}^{K}$

\ForAll{local regions $I_k$}
    \State $Y_k \gets \mathcal{M}(I_k)$
\EndFor

\If{the primary splitting direction is vertical}
    \State Merge $\{Y_k\}$ by vertical stitching with overlap deduplication
\Else
    \State Merge $\{Y_k\}$ by row-aware horizontal concatenation
\EndIf

\State Perform final integrity checking and lightweight structural repair
\State \Return final Markdown output $Y$
\end{algorithmic}
\end{algorithm}

\subsection{Discussion}

The split-then-merge strategy does not alter the core model architecture of FinixDoc-VL. Instead, it provides a lightweight engineering mechanism within the \textbf{Document Preprocessing Tool Layer} that makes ultra-large-page parsing more tractable in deployment. For standard pages, the direct-parsing path is preserved, with only minimal routing overhead before native FinixDoc-VL parsing. For oversized pages, the split branch reduces context-length pressure and memory pressure by converting a single difficult page into multiple manageable local regions. The subsequent reconstruction stage then restores page-level continuity as much as possible.

Overall, this strategy should be viewed as a practical engineering extension of FinixDoc for the \textit{Underexplored Large-Scale Zone}, rather than as a standalone parsing model. Its role is to improve usability and stability on ultra-large inputs by combining adaptive routing, structure-aware splitting, and direction-aware merging within the Document Preprocessing Tool Layer.

\section{Financial Homoglyph Vocabulary and Auxiliary Lexicons}
\label{appendix:financial_homoglyph_vocab}

To support the hard negative construction process described in Section~\ref{sec:contrastive_learning}, we build a \textbf{financial homoglyph vocabulary} that stores visually confusable characters frequently observed in real financial documents. The construction of this vocabulary relies on multi-view similarity filtering, which is supported by two foundational dictionaries: a stroke-order lexicon and a stroke-count lexicon \citep{wang2004image,levenshtein1966binary,sakoe1978dynamic}. This section details the resulting vocabulary and its underlying auxiliary lexicons.

\subsection{Financial Homoglyph Vocabulary}

We use the term \textit{homoglyph} in a broad practical sense to include both strict glyph-level lookalikes and near-homoglyph confusions that become difficult to distinguish in low-quality document images. This broader definition is particularly relevant in financial deployment, where visually small deviations in digits, field values, names, and medical or reimbursement terminology may lead to disproportionately large downstream business risk.

The vocabulary covers multiple character types, including Arabic numerals, uppercase and lowercase English letters, Roman numerals, symbol-like characters, and high-frequency Chinese characters commonly appearing in financial, insurance, medical, and reimbursement documents. Compared with generic visually similar character lists, this vocabulary is specifically tailored to the financial domain and emphasizes confusions that are both visually plausible and practically risky in downstream business scenarios.

Formally, for a character $c$, the vocabulary provides a candidate set
\[
\mathcal{H}(c) = \{h_1, h_2, \dots, h_m\},
\]
where each $h_i$ denotes a visually similar or perceptually confusable alternative that may replace $c$ during hard negative construction. These candidate sets are used to generate negative text samples by selectively substituting characters in the ground-truth annotation with their corresponding candidates. This process yields hard negatives that remain lexically close to the positive target while introducing realistic visual ambiguity.

The full vocabulary is constructed through candidate retrieval followed by similarity-based filtering; the entries below are illustrative rather than exhaustive. Table~\ref{tab:financial_homoglyph_examples} shows representative examples from the vocabulary. In implementation, the vocabulary is stored as a key--value mapping from target characters to candidate homoglyph sets. A simplified example is shown below:

\begin{verbatim}
{
    "0": ["O", "8", "6", "9", "D"],
    "I": ["l", "1", "|", "!"],
    "|": ["I", "l", "!", "丨"],
    "维": ["唯", "淮", "推", "准"],
    "清": ["靖", "情", "请", "洁"],
    "痛": ["疼", "病", "痹", "痪"]
}
\end{verbatim}

During training data construction, we sample replacements from this vocabulary under predefined rules so that the generated negative samples better match real OCR and parsing errors observed in financial deployment. In this way, the contrastive learning stage is encouraged to distinguish not only semantically different strings, but also visually confusable alternatives that are highly relevant to practical business risk.

\begin{table*}[htbp]
\centering
\scriptsize
\caption{Representative examples from the financial homoglyph vocabulary.}
\label{tab:financial_homoglyph_examples}
\begin{tabular}{lp{0.31\textwidth}|lp{0.31\textwidth}}
\toprule
Target Character & Example Homoglyphs & Target Character & Example Homoglyphs \\
\midrule
0   & O, 8, 6, 9, D                 & 维 & 唯, 淮, 推, 准 \\
O   & 0, Q, D                       & 清 & 靖, 情, 请, 洁 \\
I   & l, 1, \texttt{|}, !           & 请 & 情, 清, 谱, 讲, 证 \\
l   & I, i, \texttt{|}, !           & 组 & 细, 织, 绪, 级, 纸 \\
\texttt{|} & I, l, !, 丨            & 痛 & 疼, 病, 痹, 痪 \\
1   & I, l, 7, !                    & 靖 & 情, 晴, 清 \\
B   & 8, D, P                       & 情 & 请, 清, 青, 倩 \\
S   & 5, 8                          & 细 & 维, 组, 纸, 绪 \\
C   & G, 0, 6                       & 浩 & 洁, 活, 清 \\
D   & O, 0, B                       & 继 & 维, 续, 编, 细 \\
T   & I, 7                          & 准 & 淮, 推, 难, 维 \\
U   & V, O, D                       & 唯 & 维, 淮, 谁 \\
\bottomrule
\end{tabular}
\end{table*}

\subsection{Stroke-Order Lexicon}
\label{appendix:stroke_order_lexicon}

To compute the stroke-order similarity $S_{\mathrm{order}}(c_i, c_j)$ during candidate retrieval, we construct a stroke-order lexicon covering over 20,000 common Chinese characters. This lexicon is essential for capturing the internal composition of characters that cannot be fully resolved by image-level structural similarity (SSIM) alone \citep{wang2004image}.

As shown in the snippet below, each character is decomposed into two aligned lists:
\texttt{stroke\_type\_sequence}, which records the normalized geometric type of each stroke
using a compact stroke-code nomenclature (e.g., \texttt{S} for Shu/Vertical,
\texttt{H} for Heng/Horizontal, \texttt{P} for Pie/Left-Falling,
\texttt{D} for Dian/Dot, and \texttt{Z} for Zhe/Turning strokes, including complex
multi-turn variants), and \texttt{stroke\_name\_sequence}, which provides the corresponding
descriptive stroke name. During candidate retrieval, these sequences allow us to robustly
estimate character similarity using Dynamic Time Warping (DTW) \citep{sakoe1978dynamic} and Longest Common
Subsequence (LCS), which are resilient to localized order shifts. Additionally, alternative
stroke names (e.g., separated by slashes in \texttt{stroke\_name\_sequence}) are included
to accommodate variant naming conventions and enhance matching robustness.

A simplified snippet of the stroke-order data structure is provided below:

\begin{verbatim}
{
    "吖": {
        "stroke_type_sequence": [
            "S", "Z", "H", "D", "P", "S"
        ],
        "stroke_name_sequence": [
            "竖", "横折", "横", "点", "撇", "竖"
        ]
    },
    "阿": {
        "stroke_type_sequence": [
            "Z_multi", "S", "H", "S",
            "Z", "H", "S_hook"
        ],
        "stroke_name_sequence": [
            "横折折折钩/横撇弯钩", "竖", "横", "竖",
            "横折", "横", "竖钩"
        ]
    },
    "啊": {
        "stroke_type_sequence": [
            "S", "Z", "H", "Z_multi", "S",
            "H", "S", "Z", "H", "S_hook"
        ],
        "stroke_name_sequence": [
            "竖", "横折", "横", "横折折折钩/横撇弯钩",
            "竖", "横", "竖", "横折", "横", "竖钩"
        ]
    }
}
\end{verbatim}

\subsection{Stroke-Count Lexicon}
\label{appendix:stroke_count_lexicon}

The stroke-count lexicon provides a lightweight auxiliary cue for writing complexity, which directly supports the computation of the stroke-count similarity $S_{\mathrm{count}}(c_i, c_j)$. In practical financial document parsing, characters with identical or highly similar stroke counts (e.g., ``清'' and ``情'') are particularly susceptible to confusion under severe image degradation, such as camera blur or low optical resolution.

This lexicon maps each candidate character directly to its integer stroke count. To illustrate the distribution and coverage of this lexicon, Table~\ref{tab:stroke_count_examples} presents representative characters grouped by their respective stroke counts.

\begin{table}[htbp]
\centering
\caption{Representative examples from the stroke-count lexicon.}
\label{tab:stroke_count_examples}
\begin{tabular}{cp{10cm}}
\toprule
\textbf{Stroke Count} & \textbf{Example Characters} \\
\midrule
1 & 一, 丨, 丶, 丿, 乙 \\
2 & 丁, 七, 乃, 九, 了, 二, 乜 \\
3 & 万, 丈, 三, 上, 下, 久, 义, 亡, 个, 丫 \\
4 & 不, 丑, 专, 中, 丰, 乌, 书, 云, 互, 五 \\
5 & 且, 世, 丘, 业, 东, 丝, 乎, 乏, 乐, 主 \\
6 & 丞, 丢, 两, 争, 亘, 亚, 交, 亥, 亦, 产 \\
\bottomrule
\end{tabular}
\end{table}

\newpage
\section{Qualitative Comparison Results on FinixDigital and FinixPhoto}
\label{appendix:qualitative_comparison}

This appendix provides qualitative comparison results of \textbf{FinixDoc} against the best specialized model and the best general-purpose VLM selected for each of the \textbf{FinixDigital} and \textbf{FinixPhoto} evaluation tracks. For each subset, we select representative samples that highlight the typical structural, lexical, and robustness challenges of real-world financial document parsing.

The comparison is organized along two evaluation axes that match the design of the two tracks:
\begin{itemize}
    \item \textbf{FinixDigital} (financial digitally native documents): we compare against \textbf{MinerU~2.5} as the best specialized model and \textbf{Qwen3-VL-235B-A22B-Instruct} as the best general-purpose VLM. The selected samples emphasize dense table structures, deep heading hierarchies, and domain-specific formatting such as superscript footnote references.

    \item \textbf{FinixPhoto} (camera-captured documents): we compare against \textbf{Youtu-Parsing} as the best specialized model and \textbf{Kimi-K2.5} as the best general-purpose VLM. The selected samples emphasize OCR-level robustness under motion blur, perspective distortion, and printing artifacts, as well as the tendency of competing models to hallucinate content beyond the visible image.
\end{itemize}

As illustrated in Figures~\ref{fig:finixdigital-1}--\ref{fig:finixphoto-3}, each example uses a four-way side-by-side layout: the original page (or per-field overlay), the FinixDoc output, the best specialized model output, and the best general-purpose model output. The primary purpose of this layout is to enable a direct comparison of structural fidelity, content accuracy, and robustness across different systems under the same input conditions.

\subsection{FinixDigital: Comparison with MinerU~2.5 and Qwen3-VL-235B-A22B-Instruct}
\label{appendix:finixdigital}

\begin{figure}[p]
    \centering
    \includegraphics[width=\textwidth,height=0.88\textheight,keepaspectratio]{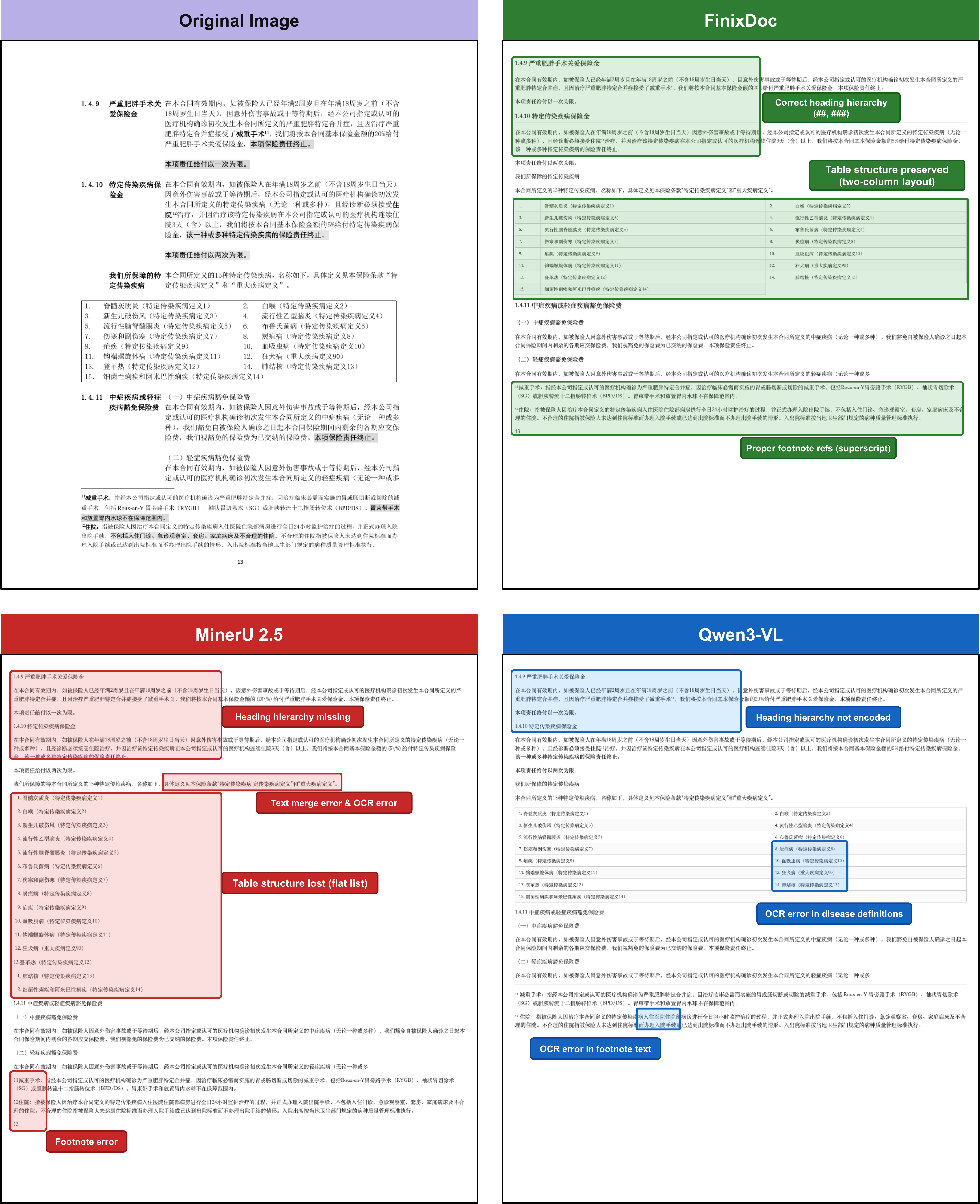}
    \caption{\small\textbf{FinixDigital Sample~1.} Comparison of parsing results for an insurance clause page containing a two-column disease list table and footnote references. FinixDoc correctly preserves the two-column table structure for the 15 infectious diseases, maintains proper heading hierarchy (\texttt{\#\#}, \texttt{\#\#\#}), and retains superscript footnote references ($^{11}$, $^{12}$). In contrast, MinerU~2.5 collapses the table into a flat list, fails to encode the heading hierarchy, merges neighboring text around the disease-list introduction, and introduces OCR errors. Qwen3-VL-235B-A22B-Instruct preserves the table layout more completely than MinerU~2.5, but still fails to encode the heading hierarchy, changes domain terms in the disease-list entries (e.g., \emph{特定传染疾病定义} becomes \emph{特定传染病定义}), and introduces a footnote-level OCR error (e.g., \emph{入住医院} becomes \emph{入往医院}).}
    \label{fig:finixdigital-1}
\end{figure}

\begin{figure}[p]
    \centering
    \includegraphics[width=\textwidth,height=0.88\textheight,keepaspectratio]{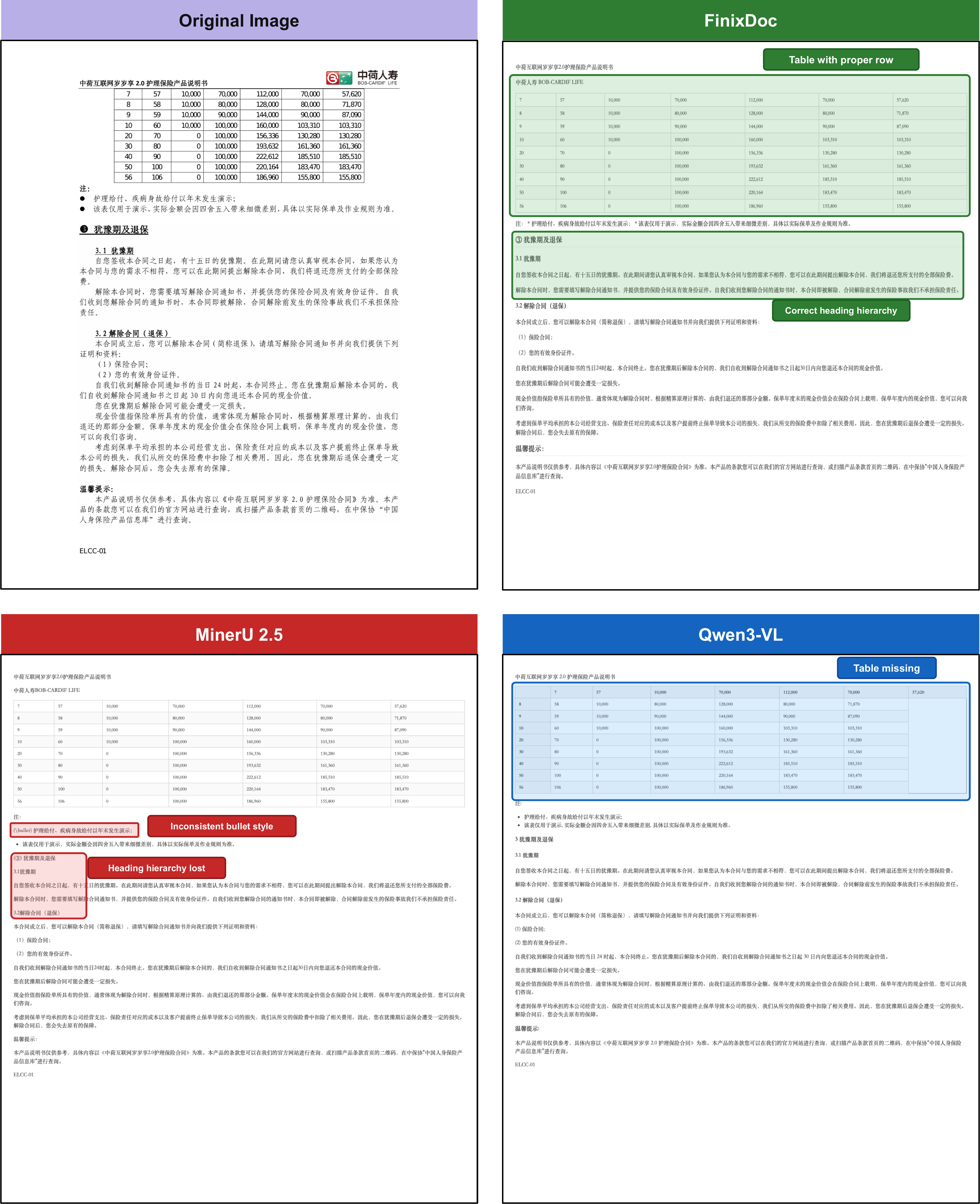}
    \caption{\textbf{FinixDigital Sample~2.} Comparison of parsing results for an insurance product specification page containing a financial illustration table and section headings. FinixDoc preserves the numeric table grid, reconstructs the heading hierarchy (\texttt{\#\#\#~3}, \texttt{\#\#\#\#~3.1}, \texttt{\#\#\#\#~3.2}), and maintains consistent list formatting with \texttt{(1)} and \texttt{(2)} numbering. MinerU~2.5 fails to render the circled section marker and subsection hierarchy as Markdown headings and mixes bullet styles in the notes. Qwen3-VL-235B-A22B-Instruct misinterprets the first numeric data row as a table header and introduces an HTML/dataframe-style table with an extra index-like column.}
    \label{fig:finixdigital-2}
\end{figure}

\begin{figure}[p]
    \centering
    \includegraphics[width=\textwidth,height=0.88\textheight,keepaspectratio]{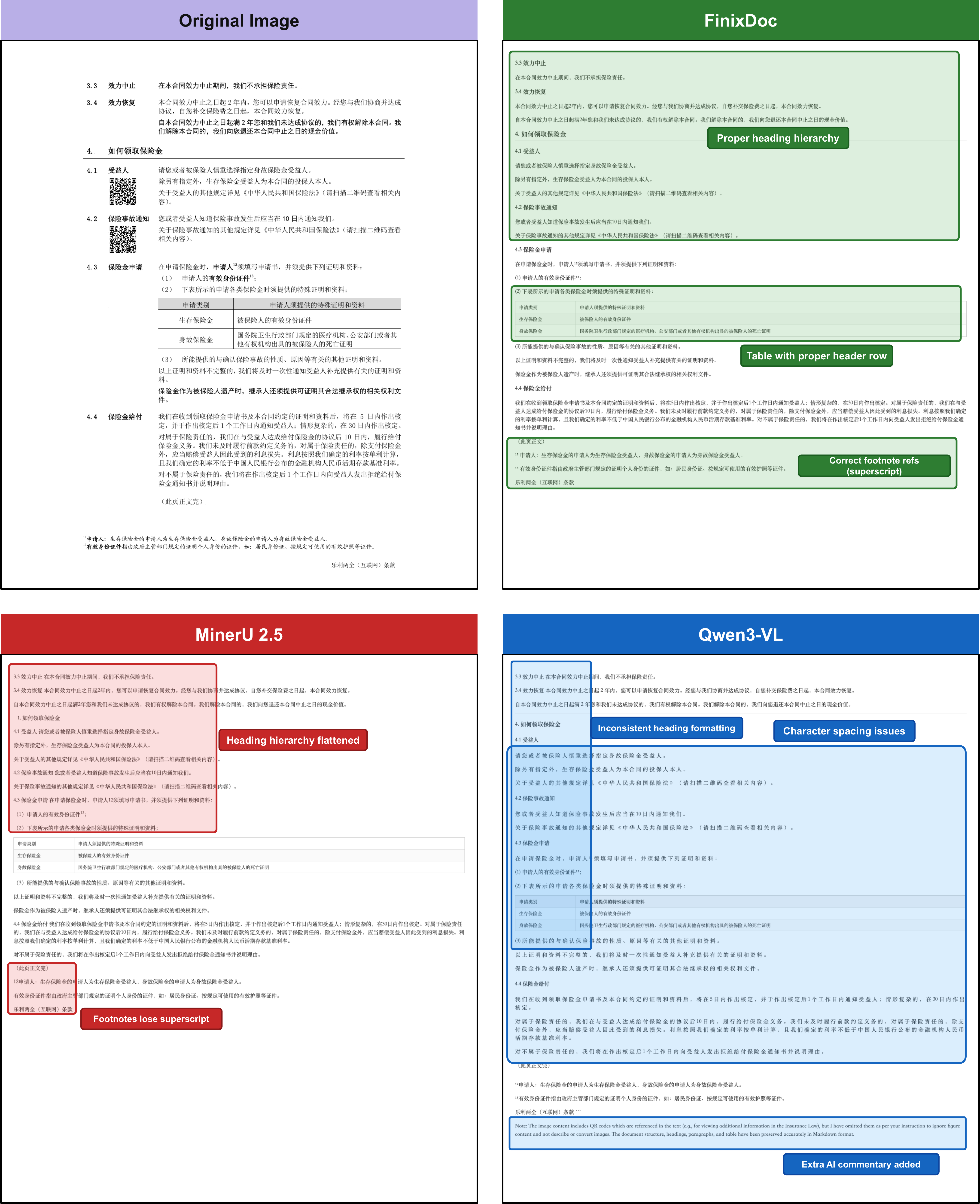}
    \caption{\textbf{FinixDigital Sample~3.} Comparison of parsing results for an insurance contract page with nested headings, a structured table, and footnote references. FinixDoc accurately reconstructs the heading hierarchy (\texttt{\#\#\#~3.3}, \texttt{\#\#\#~3.4}, \texttt{\#\#~4}, \texttt{\#\#\#~4.1}--\texttt{\#\#\#~4.4}), preserves the table with proper headers, and retains superscript footnote references ($^{12}$, $^{13}$). MinerU~2.5 flattens the heading hierarchy, loses or inconsistently renders superscript footnotes (e.g., $^{12}$ becomes plain text), and degrades the page-ending footnote area. Qwen3-VL-235B-A22B-Instruct preserves the table but inserts spurious spacing between Chinese characters, appends extraneous AI commentary after the parsed page, and shows inconsistent heading formatting.}
    \label{fig:finixdigital-3}
\end{figure}

Figures~\ref{fig:finixdigital-1}--\ref{fig:finixdigital-3} showcase FinixDoc's performance on digitally native financial documents from the FinixDigital track. Each figure presents a side-by-side comparison of the rendered Markdown output: the original page image, FinixDoc's parsed and rendered result, the result of the best specialized model (MinerU~2.5), and the result of the best general-purpose VLM (Qwen3-VL-235B-A22B-Instruct).

\clearpage

\subsection{FinixPhoto: Comparison with Youtu-Parsing and Kimi-K2.5}
\label{appendix:finixphoto}

\begin{figure}[H]
    \centering
    \includegraphics[width=\textwidth,height=0.88\textheight,keepaspectratio]{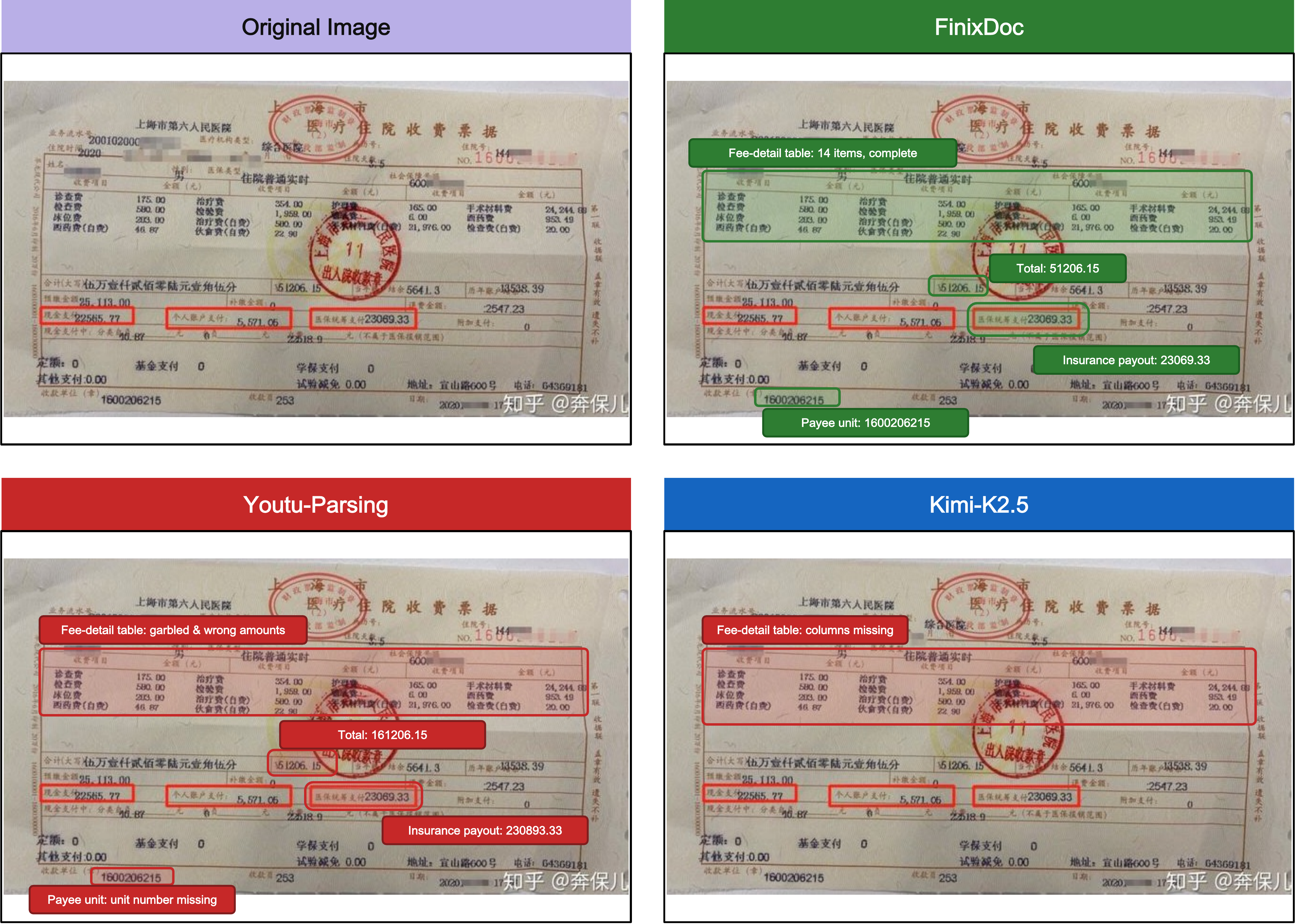}
    \caption{\small\textbf{FinixPhoto Sample~1.} Per-field comparison on a Shanghai inpatient receipt captured under challenging conditions. Four fields are highlighted: the fee-detail table, the total amount (\texttt{51206.15}), the medical-insurance payout (\texttt{23069.33}), and the payee-unit number (\texttt{1600206215}). FinixDoc parses all four correctly. Youtu-Parsing reads the total as \texttt{161206.15}, the insurance payout as \texttt{230893.33}, drops the payee-unit number, and produces a heavily garbled fee table. Kimi-K2.5 gets the numeric fields right but its fee table is missing most of the 14 items present in the original.}
    \label{fig:finixphoto-1}
\end{figure}

\begin{figure}[p]
    \centering
    \includegraphics[width=\textwidth,height=0.88\textheight,keepaspectratio]{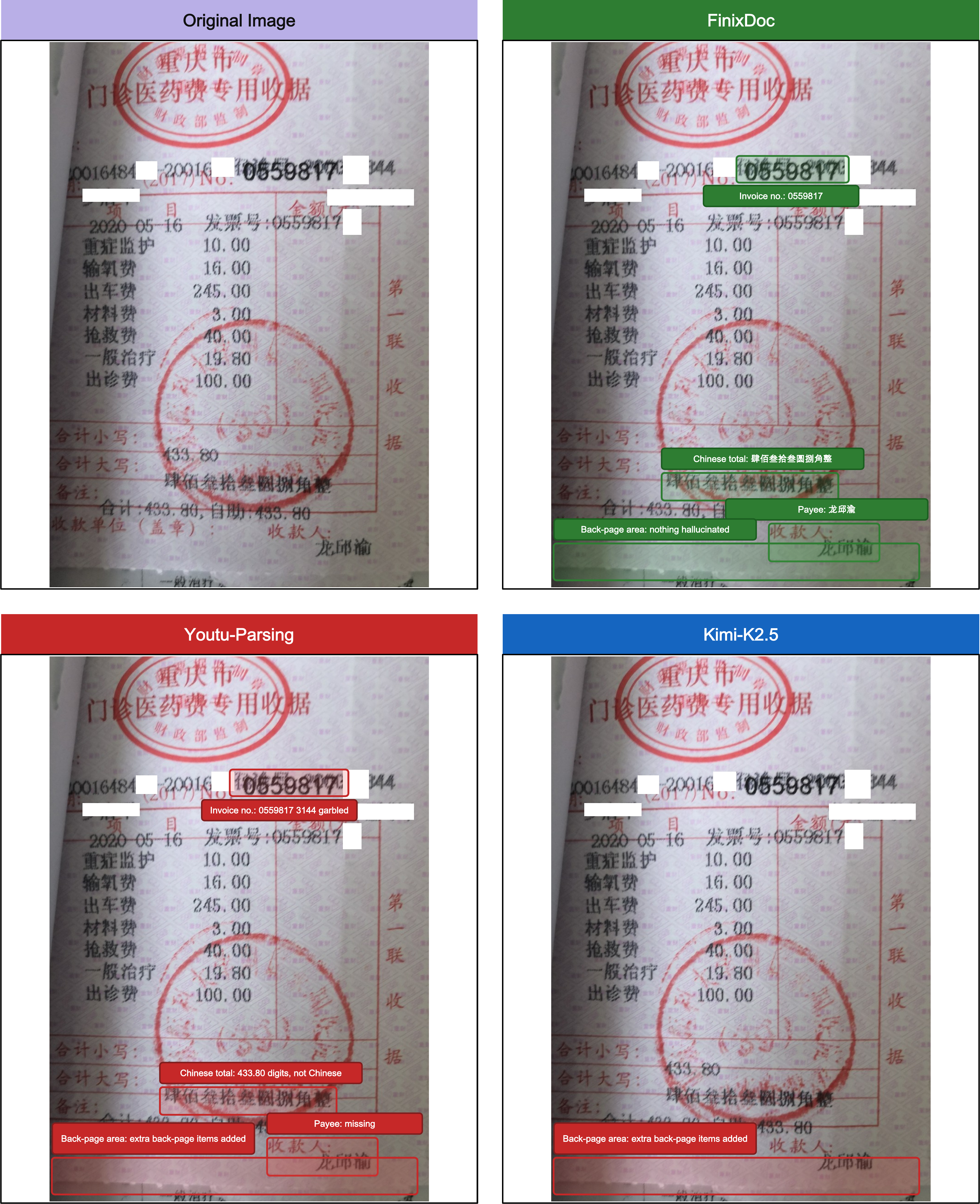}
    \caption{\small\textbf{FinixPhoto Sample~2.} Per-field comparison on a Chongqing outpatient medical receipt. Four fields are highlighted: the invoice number (\texttt{0559817}), the Chinese-capital total (肆佰叁拾叁圆捌角整), the payee (龙邱渝), and a synthetic ``back-page area'' box at the bottom strip of the receipt. FinixDoc gets the invoice number, the Chinese total, and the payee correct, and crucially does not hallucinate any content beyond what is visible on the front of the receipt. Youtu-Parsing renders the Chinese total as a plain ``\texttt{433.80}'' and drops the payee. Both Youtu-Parsing and Kimi-K2.5 hallucinate extra back-page items (e.g., CT, 重症监护, 中医治疗) that are not visible in the input image.}
    \label{fig:finixphoto-2}
\end{figure}

\begin{figure}[H]
    \centering
    \includegraphics[width=\textwidth,height=0.88\textheight,keepaspectratio]{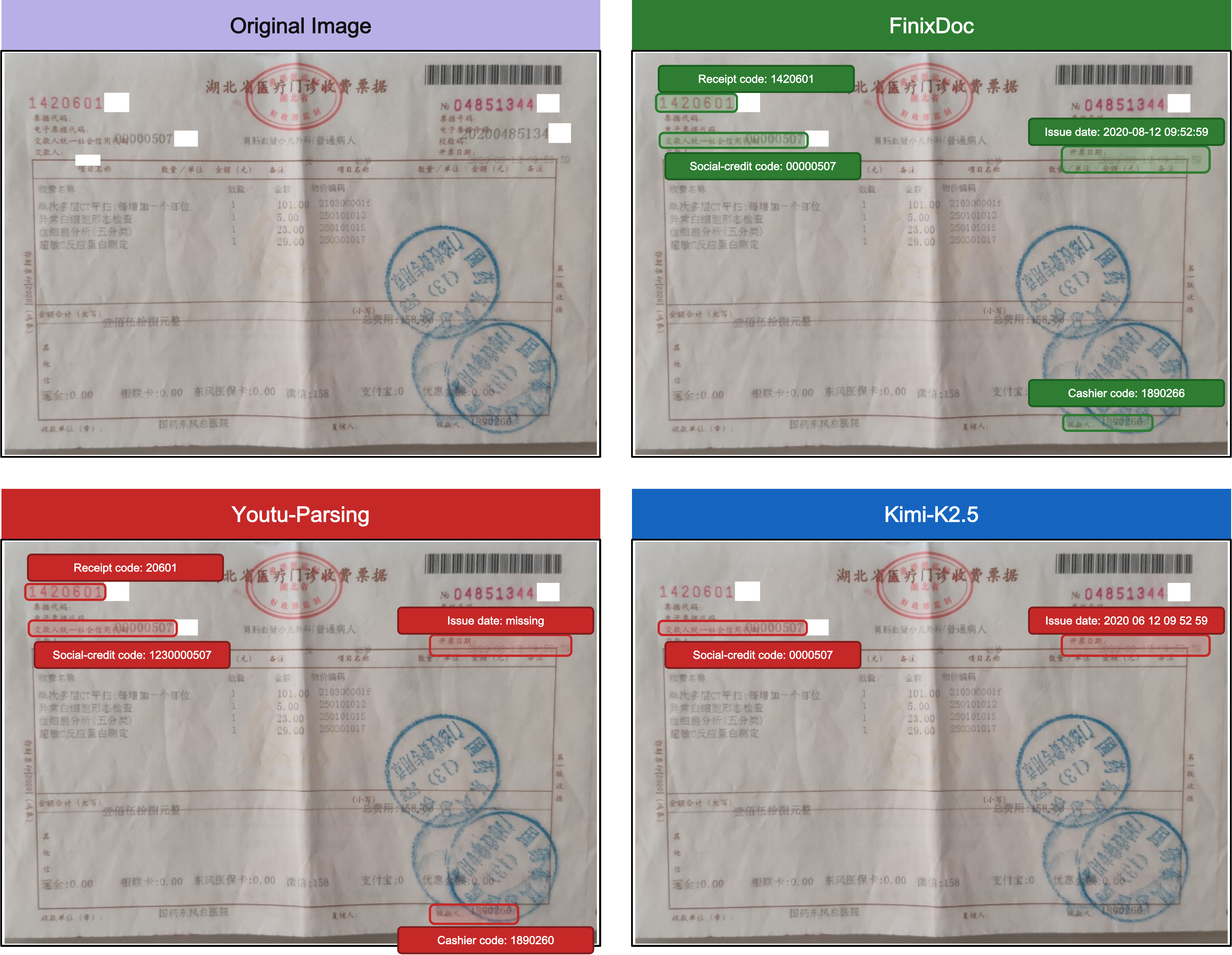}
    \caption{\small\textbf{FinixPhoto Sample~3.} Per-field comparison on a Hubei outpatient medical invoice. Four fields are highlighted: the receipt code (\texttt{1420601}), the unified social-credit code (\texttt{00000507}), the issue date (\texttt{2020-08-12 09:52:59}), and the cashier code (\texttt{1890266}). FinixDoc gets all four correct. Youtu-Parsing truncates the receipt code to \texttt{20601}, prefixes the social-credit code with an extraneous \texttt{123}, drops the issue date, and reads the cashier code as \texttt{1890260}. Kimi-K2.5 reads \texttt{0000507} (one leading zero missing) for the social-credit code and \texttt{2020 06 12 09 52 59} (wrong month, lost separators) for the issue date.}
    \label{fig:finixphoto-3}
\end{figure}

Figures~\ref{fig:finixphoto-1}--\ref{fig:finixphoto-3} illustrate FinixDoc's robustness on camera-captured document images from the FinixPhoto track. Each figure is organized as a $2{\times}2$ grid composed of (i) the unannotated original image, and (ii)--(iv) the same original image overlaid with the ground-truth bounding boxes of selected fields, color-coded by whether each model parsed that field correctly (green) or incorrectly (red). For incorrect fields, the model's actual output is rendered next to the corresponding box so that the reader can see the exact error mode. All bounding boxes are taken from the FinixPhoto ground-truth annotations.

%% file: tool/biblio.bib
@inproceedings{smith2007tesseract,
  title={An Overview of the {Tesseract OCR} Engine},
  author={Smith, Ray},
  booktitle={Ninth International Conference on Document Analysis and Recognition},
  volume={2},
  pages={629--633},
  year={2007},
  organization={IEEE}
}

@article{zhang2024documentparsingunveiled,
  title={Document Parsing Unveiled: Techniques, Challenges, and Prospects for Structured Information Extraction},
  author={Zhang, Qintong and Wang, Bin and Huang, Victor Shea-Jay and Zhang, Junyuan and Wang, Zhengren and Liang, Hao and He, Conghui and Zhang, Wentao},
  journal={arXiv preprint arXiv:2410.21169},
  year={2024}
}

@article{levenshtein1966binary,
  title={Binary Codes Capable of Correcting Deletions, Insertions, and Reversals},
  author={Levenshtein, Vladimir I.},
  journal={Soviet Physics Doklady},
  volume={10},
  number={8},
  pages={707--710},
  year={1966}
}

@article{sakoe1978dynamic,
  title={Dynamic Programming Algorithm Optimization for Spoken Word Recognition},
  author={Sakoe, Hiroaki and Chiba, Seibi},
  journal={IEEE Transactions on Acoustics, Speech, and Signal Processing},
  volume={26},
  number={1},
  pages={43--49},
  year={1978},
  publisher={IEEE}
}

@article{zhang1989simple,
  title={Simple Fast Algorithms for the Editing Distance Between Trees and Related Problems},
  author={Zhang, Kaizhong and Shasha, Dennis},
  journal={SIAM Journal on Computing},
  volume={18},
  number={6},
  pages={1245--1262},
  year={1989},
  publisher={SIAM}
}

@article{wang2004image,
  title={Image Quality Assessment: From Error Visibility to Structural Similarity},
  author={Wang, Zhou and Bovik, Alan C. and Sheikh, Hamid R. and Simoncelli, Eero P.},
  journal={IEEE Transactions on Image Processing},
  volume={13},
  number={4},
  pages={600--612},
  year={2004},
  publisher={IEEE}
}

@inproceedings{schroff2015facenet,
  title={{FaceNet}: A Unified Embedding for Face Recognition and Clustering},
  author={Schroff, Florian and Kalenichenko, Dmitry and Philbin, James},
  booktitle={Proceedings of the IEEE Conference on Computer Vision and Pattern Recognition},
  pages={815--823},
  year={2015}
}

@article{schulman2017proximal,
  title={Proximal Policy Optimization Algorithms},
  author={Schulman, John and Wolski, Filip and Dhariwal, Prafulla and Radford, Alec and Klimov, Oleg},
  journal={arXiv preprint arXiv:1707.06347},
  year={2017}
}

@inproceedings{khosla2020supervised,
  title={Supervised Contrastive Learning},
  author={Khosla, Prannay and Teterwak, Piotr and Wang, Chen and Sarna, Aaron and Tian, Yonglong and Isola, Phillip and Maschinot, Aaron and Liu, Ce and Krishnan, Dilip},
  booktitle={Advances in Neural Information Processing Systems},
  volume={33},
  pages={18661--18673},
  year={2020}
}

@inproceedings{radford2021learning,
  title={Learning Transferable Visual Models From Natural Language Supervision},
  author={Radford, Alec and Kim, Jong Wook and Hallacy, Chris and Ramesh, Aditya and Goh, Gabriel and Agarwal, Sandhini and Sastry, Girish and Askell, Amanda and Mishkin, Pamela and Clark, Jack and Krueger, Gretchen and Sutskever, Ilya},
  booktitle={Proceedings of the 38th International Conference on Machine Learning},
  pages={8748--8763},
  year={2021}
}

@inproceedings{dosovitskiy2021image,
  title={An Image is Worth 16x16 Words: Transformers for Image Recognition at Scale},
  author={Dosovitskiy, Alexey and Beyer, Lucas and Kolesnikov, Alexander and Weissenborn, Dirk and Zhai, Xiaohua and Unterthiner, Thomas and Dehghani, Mostafa and Minderer, Matthias and Heigold, Georg and Gelly, Sylvain and Uszkoreit, Jakob and Houlsby, Neil},
  booktitle={International Conference on Learning Representations},
  year={2021}
}

@inproceedings{ouyang2022training,
  title={Training Language Models to Follow Instructions with Human Feedback},
  author={Ouyang, Long and Wu, Jeffrey and Jiang, Xu and Almeida, Diogo and Wainwright, Carroll and Mishkin, Pamela and Zhang, Chong and Agarwal, Sandhini and Slama, Katarina and Ray, Alex and others},
  booktitle={Advances in Neural Information Processing Systems},
  volume={35},
  pages={27730--27744},
  year={2022}
}

@article{shao2024deepseekmath,
  title={{DeepSeekMath}: Pushing the Limits of Mathematical Reasoning in Open Language Models},
  author={Shao, Zhihong and Wang, Peiyi and Zhu, Qihao and Xu, Runxin and Song, Junxiao and Bi, Xiao and Zhang, Haowei and Zhang, Mingchuan and Li, Y. K. and Wu, Y. and Guo, Daya},
  journal={arXiv preprint arXiv:2402.03300},
  year={2024}
}

@article{guo2025deepseekr1,
  title={{DeepSeek-R1}: Incentivizing Reasoning Capability in {LLMs} via Reinforcement Learning},
  author={{DeepSeek-AI} and Guo, Daya and Yang, Dejian and Zhang, Haowei and Song, Junxiao and Wang, Peiyi and Zhu, Qihao and Xu, Runxin and Zhang, Ruoyu and Ma, Shirong and Bi, Xiao and others},
  journal={arXiv preprint arXiv:2501.12948},
  year={2025}
}

@inproceedings{yao2023react,
  title={{ReAct}: Synergizing Reasoning and Acting in Language Models},
  author={Yao, Shunyu and Zhao, Jeffrey and Yu, Dian and Du, Nan and Shafran, Izhak and Narasimhan, Karthik R. and Cao, Yuan},
  booktitle={The Eleventh International Conference on Learning Representations},
  year={2023}
}

@inproceedings{schick2023toolformer,
  title={{Toolformer}: Language Models Can Teach Themselves to Use Tools},
  author={Schick, Timo and Dwivedi-Yu, Jane and Dessi, Roberto and Raileanu, Roberta and Lomeli, Maria and Hambro, Eric and Zettlemoyer, Luke and Cancedda, Nicola and Scialom, Thomas},
  booktitle={Advances in Neural Information Processing Systems},
  volume={36},
  pages={68539--68551},
  year={2023}
}

@inproceedings{li2023evaluatingpope,
  title={Evaluating Object Hallucination in Large Vision-Language Models},
  author={Li, Yifan and Du, Yifan and Zhou, Kun and Wang, Jinpeng and Zhao, Wayne Xin and Wen, Ji-Rong},
  booktitle={Proceedings of the 2023 Conference on Empirical Methods in Natural Language Processing},
  pages={292--305},
  year={2023}
}

@article{liu2024hallucinationsurvey,
  title={A Survey on Hallucination in Large Vision-Language Models},
  author={Liu, Hanchao and Xue, Wenyuan and Chen, Yifei and Chen, Dapeng and Zhao, Xiutian and Wang, Ke and Hou, Liping and Li, Rongjun and Peng, Wei},
  journal={arXiv preprint arXiv:2402.00253},
  year={2024}
}

@inproceedings{kim2022donut,
  title={{OCR}-Free Document Understanding Transformer},
  author={Kim, Geewook and Hong, Teakgyu and Yim, Moonbin and Nam, JeongYeon and Park, Jinyoung and Yim, Jinyeong and Hwang, Wonseok and Yun, Sangdoo and Han, Dongyoon and Park, Seunghyun},
  booktitle={European Conference on Computer Vision},
  pages={498--517},
  year={2022},
  organization={Springer}
}

@inproceedings{zhong2019publaynet,
  title={{PubLayNet}: Largest Dataset Ever for Document Layout Analysis},
  author={Zhong, Xu and Tang, Jianbin and Yepes, Antonio Jimeno},
  booktitle={2019 International Conference on Document Analysis and Recognition},
  pages={1015--1022},
  year={2019},
  organization={IEEE}
}

@inproceedings{li2020docbank,
  title={{DocBank}: A Benchmark Dataset for Document Layout Analysis},
  author={Li, Minghao and Xu, Yiheng and Cui, Lei and Huang, Shaohan and Wei, Furu and Li, Zhoujun and Zhou, Ming},
  booktitle={Proceedings of the 28th International Conference on Computational Linguistics},
  pages={949--960},
  year={2020}
}

@inproceedings{zhong2020image,
  title={Image-Based Table Recognition: Data, Model, and Evaluation},
  author={Zhong, Xu and ShafieiBavani, Elaheh and Jimeno Yepes, Antonio},
  booktitle={European Conference on Computer Vision},
  pages={564--580},
  year={2020},
  organization={Springer}
}

@inproceedings{pfitzmann2022doclaynet,
  title={{DocLayNet}: A Large Human-Annotated Dataset for Document-Layout Segmentation},
  author={Pfitzmann, Birgit and Auer, Christoph and Dolfi, Michele and Nassar, Ahmed S. and Staar, Peter W. J.},
  booktitle={Proceedings of the 28th ACM SIGKDD Conference on Knowledge Discovery and Data Mining},
  pages={3743--3751},
  year={2022}
}

@article{wang2025infinityparser,
  title={{Infinity Parser}: Layout Aware Reinforcement Learning for Scanned Document Parsing},
  author={Wang, Baode and Wu, Biao and Li, Weizhen and Fang, Meng and Huang, Zuming and Huang, Jun and Wang, Haozhe and Liang, Yanjie and Chen, Ling and Chu, Wei and Qi, Yuan},
  journal={arXiv preprint arXiv:2506.03197},
  year={2025}
}

@inproceedings{lee2023pix2struct,
  title={{Pix2Struct}: Screenshot Parsing as Pretraining for Visual Language Understanding},
  author={Lee, Kenton and Joshi, Mandar and Turc, Iulia Raluca and Hu, Hexiang and Liu, Fangyu and Eisenschlos, Julian Martin and Khandelwal, Urvashi and Shaw, Peter and Chang, Ming-Wei and Toutanova, Kristina},
  booktitle={Proceedings of the 40th International Conference on Machine Learning},
  pages={18893--18912},
  year={2023}
}

@article{blecher2023nougat,
  title={{Nougat}: Neural Optical Understanding for Academic Documents},
  author={Blecher, Lukas and Cucurull, Guillem and Scialom, Thomas and Stojnic, Robert},
  journal={arXiv preprint arXiv:2308.13418},
  year={2023}
}

@inproceedings{ouyang2025omnidocbench,
  title={{OmniDocBench}: Benchmarking Diverse {PDF} Document Parsing with Comprehensive Annotations},
  author={Ouyang, Linke and Qu, Yuan and Zhou, Hongbin and Zhu, Jiawei and Zhang, Rui and Lin, Qunshu and Wang, Bin and Zhao, Zhiyuan and Jiang, Man and Zhao, Xiaomeng and Shi, Jin and Wu, Fan and Chu, Pei and Liu, Minghao and Li, Zhenxiang and Xu, Chao and Zhang, Bo and Shi, Botian and Tu, Zhongying and He, Conghui},
  booktitle={2025 IEEE/CVF Conference on Computer Vision and Pattern Recognition (CVPR)},
  pages={24838--24848},
  year={2025}
}

@article{bai2025qwen25vl,
  title={{Qwen2.5-VL} Technical Report},
  author={Bai, Shuai and Chen, Keqin and Liu, Xuejing and Wang, Jialin and Ge, Wenbin and Song, Sibo and Dang, Kai and Wang, Peng and Wang, Shijie and Tang, Jun and others},
  journal={arXiv preprint arXiv:2502.13923},
  year={2025}
}

@article{qwen2025qwen3vl,
  title={{Qwen3-VL} Technical Report},
  author={{Qwen Team}},
  journal={arXiv preprint arXiv:2511.21631},
  year={2025}
}

@article{kimiteam2025kimivl,
  title={{Kimi-VL} Technical Report},
  author={{Kimi Team} and Du, Angang and Yin, Bohong and Xing, Bowei and Qu, Bowen and Wang, Bowen and others},
  journal={arXiv preprint arXiv:2504.07491},
  year={2025}
}

@article{kimiteam2026k25,
  title={{Kimi K2.5}: Visual Agentic Intelligence},
  author={{Kimi Team}},
  journal={arXiv preprint arXiv:2602.02276},
  year={2026}
}

@misc{qwen2026qwen35blog,
  title={{Qwen3.5}: Towards Native Multimodal Agents},
  author={{Qwen Team}},
  howpublished={\url{https://qwen.ai/blog?id=qwen3.5}},
  year={2026},
  note={Official release blog}
}

@article{paddleocr2025technical,
  title={{PaddleOCR} 3.0 Technical Report},
  author={Cui, Cheng and Sun, Ting and Lin, Manhui and Gao, Tingquan and Zhang, Yubo and Liu, Jiaxuan and Wang, Xueqing and Zhang, Zelun and Zhou, Changda and Liu, Hongen and Zhang, Yue and Lv, Wenyu and Huang, Kui and Zhang, Yichao and Zhang, Jing and Zhang, Jun and Liu, Yi and Yu, Dianhai and Ma, Yanjun},
  journal={arXiv preprint arXiv:2507.05595},
  year={2025}
}

@article{sun2025ppdoclayout,
  title={{PP-DocLayout}: A Unified Document Layout Detection Model to Accelerate Large-Scale Data Construction},
  author={Sun, Ting and Cui, Cheng and Du, Yuning and Liu, Yi},
  journal={arXiv preprint arXiv:2503.17213},
  year={2025}
}

@article{cui2025paddleocrvl,
  title={{PaddleOCR-VL}: Boosting Multilingual Document Parsing via a 0.9B Ultra-Compact Vision-Language Model},
  author={Cui, Cheng and Sun, Ting and Liang, Suyin and Gao, Tingquan and Zhang, Zelun and Liu, Jiaxuan and Wang, Xueqing and Zhou, Changda and Liu, Hongen and Lin, Manhui and others},
  journal={arXiv preprint arXiv:2510.14528},
  year={2025}
}

@article{cui2026paddleocrvl15,
  title={{PaddleOCR-VL-1.5}: Towards a Multi-Task 0.9B {VLM} for Robust In-the-Wild Document Parsing},
  author={Cui, Cheng and Sun, Ting and Liang, Suyin and Gao, Tingquan and Zhang, Zelun and Liu, Jiaxuan and Wang, Xueqing and Zhou, Changda and Liu, Hongen and Lin, Manhui and Zhang, Yue and Zhang, Yubo and Liu, Yi and Yu, Dianhai and Ma, Yanjun},
  journal={arXiv preprint arXiv:2601.21957},
  year={2026}
}

@article{wei2026deepseekocr2,
  title={{DeepSeek-OCR 2}: Visual Causal Flow},
  author={Wei, Haoran and Sun, Yaofeng and Li, Yukun},
  journal={arXiv preprint arXiv:2601.20552},
  year={2026}
}

@article{glm2025vl,
  title={{GLM-4.5V} and {GLM-4.1V-Thinking}: Towards Versatile Multimodal Reasoning with Scalable Reinforcement Learning},
  author={{V Team} and Hong, Wenyi and Yu, Wenmeng and Gu, Xiaotao and Wang, Guo and Gan, Guobing and Tang, Haomiao and Cheng, Jiale and Qi, Ji and Ji, Junhui and Pan, Lihang and Duan, Shuaiqi and others},
  journal={arXiv preprint arXiv:2507.01006},
  year={2025}
}

@article{duan2026glmocr,
  title={{GLM-OCR} Technical Report},
  author={Duan, Shuaiqi and Xue, Yadong and Wang, Weihan and Su, Zhe and Liu, Huan and Yang, Sheng and Gan, Guobing and Wang, Guo and Wang, Zihan and Yan, Shengdong and others},
  journal={arXiv preprint arXiv:2603.10910},
  year={2026}
}

@article{dotsocr2025,
  title={{dots.ocr}: Multilingual Document Layout Parsing in a Single Vision-Language Model},
  author={Li, Yumeng and Yang, Guang and Liu, Hao and Wang, Bowen and Zhang, Colin},
  journal={arXiv preprint arXiv:2512.02498},
  year={2025}
}

@article{fireredteam2026fireredocr,
  title={{FireRed-OCR} Technical Report},
  author={{FireRed Team}},
  journal={arXiv preprint arXiv:2603.01840},
  year={2026}
}

@article{youtu2026parsing,
  title={{Youtu-Parsing}: Perception, Structuring and Recognition via High-Parallelism Decoding},
  author={{Youtu-Parsing Team}},
  journal={arXiv preprint arXiv:2601.20430},
  year={2026}
}

@article{niu2025mineru25,
  title={{MinerU2.5}: A Decoupled Vision-Language Model for Efficient High-Resolution Document Parsing},
  author={Niu, Junbo and Liu, Zheng and Gu, Zhuangcheng and Wang, Bin and Ouyang, Linke and Zhao, Zhiyuan and Chu, Tao and He, Tianyao and Wu, Fan and Zhang, Qintong and others},
  journal={arXiv preprint arXiv:2509.22186},
  year={2025}
}
